\documentclass[twoside]{article}

\usepackage[accepted]{aistats2020}

\usepackage{amsmath}
\usepackage{mathrsfs}
\usepackage{amssymb}
\usepackage{amsfonts}
\usepackage{amsthm}
\usepackage{stmaryrd}
\usepackage{xcolor}
\usepackage{mathrsfs,dsfont}
\usepackage{bbm}
\usepackage[round]{natbib}

\usepackage{hyperref}
\usepackage{enumerate}
\usepackage{mathtools}
\usepackage{etextools}
\usepackage[capitalise,noabbrev]{cleveref}  %
\crefname{appsec}{Appendix}{Appendices}
\crefformat{equation}{(#2#1#3)}

\hypersetup{
    colorlinks=true,
    linkcolor=blue,
    urlcolor=blue,
    citecolor=blue
}

\DeclarePairedDelimiter\norm{\lVert}{\rVert}
\DeclarePairedDelimiterX{\dprod}[2]{\langle}{\rangle}{#1, #2}

\SetSymbolFont{stmry}{bold}{U}{stmry}{m}{n}

\allowdisplaybreaks

\newcommand{\iid}{\overset{\mathrm{iid}}{\sim}}

\begin{document}

\twocolumn[

\aistatstitle{Stable behaviour of infinitely wide deep neural networks}

\aistatsauthor{
    Stefano Favaro\\
    \texttt{stefano.favaro@unito.it}
     \And
     Sandra Fortini\\
     \texttt{sandra.fortini@unibocconi.it}
     \And
  Stefano Peluchetti\\
    \texttt{speluchetti@cogent.co.jp}
}

\aistatsaddress{
     Department ESOMAS\\  University of Torino\\ and Collegio Carlo Alberto\\
     \And
     Department of Decision Sciences\\  
     Bocconi University\\
     \And
     Cogent Labs\\
} ]

\begin{abstract}
We consider fully connected feed-forward deep neural networks (NNs) where weights and biases are independent and identically distributed as symmetric centered stable distributions. Then, we show that the infinite wide limit of the NN, under suitable scaling on the weights, is a stochastic process whose finite-dimensional distributions are multivariate stable distributions. The limiting process is referred to as the stable process, and it generalizes the class of Gaussian processes recently obtained as infinite wide limits of NNs \citep{matthews2018gaussian}. Parameters of the stable process can be computed via an explicit recursion over the layers of the network. Our result contributes to the theory of fully connected feed-forward deep NNs, and it paves the way to expand recent lines of research that rely on Gaussian infinite wide limits.
\end{abstract}

\section{Introduction}\label{sec:intro}

The connection between infinitely wide deep feed-forward neural networks (NNs), whose parameters at initialization are independent and identically distributed (iid) as scaled and centered Gaussian distributions, and Gaussian processes (GPs) is well known \citep{neal1995bayesian,der2006beyond,lee2018deep,g.2018gaussian,matthews2018gaussian,1yang2019scaling}. Recently, this intriguing connection has been exploited in many exciting research directions, including: i) Bayesian inference for GPs arising from infinitely wide networks \citep{lee2018deep,garriga-alonso2018deep}; ii) kernel regression for infinitely wide networks which are trained with continuous-time gradient descent via the neural tangent kernel \citep{jacot2018neural,lee2019wide,arora2019exact}; iii) analysis of the properties of infinitely wide networks as function of depth via the information propagation framework \citep{poole2016exponential,schoenholz2017deep,hayou2019impact}.
It has been shown a substantial gap between finite NNs and their corresponding infinite (wide) GPs counterparts in terms of empirical performance, at least on some of the standard benchmarks applications. Moreover, it has been shown to be a difficult task to avoid undesirable empirical properties arising in the context of very deep networks. Given that, there exists an increasing interest in expanding GPs arising in the limit of infinitely wide NNs as a way forward to close, or even reverse, this empirical performance gap and to avoid, or slow down, pathological behaviors in very deep NN.

Let $\mathcal{N}(\mu,\sigma^{2})$ denote the Gaussian distribution with mean $\mu\in\mathbb{R}$ and variance $\sigma^{2}\in\mathbb{R}^{+}$. Following the celebrated work of \cite{neal1995bayesian}, we consider the shallow NN
\begin{align*}
    f_{i}^{(1)}(x)&=\sum_{j=1}^{I}w_{i,j}^{(1)}x_{j}+b_{i}^{(1)}\\
    f_{i}^{(2)}(x,n)&=\frac{1}{\sqrt{n}}\sum_{j=1}^{n}w_{i,j}^{(2)}\phi(f_{j}^{(1)}(x))+b_{i}^{(2)},
\end{align*}
where $\phi$ is a nonlinearity, $i=1,\dots,n$, $w_{i,j}^{(1)},w_{i,j}^{(2)} \iid \mathcal{N}(0, \sigma_w^2)$, $b_{i}^{(1)},b_{i}^{(2)} \iid \mathcal{N}(0, \sigma_b^2)$ and $x \in \mathbb{R}^I$ is the input. It follows that
\begin{align*}
    f_{i}^{(1)}(x) &\iid \mathcal{N}\left(0, \sigma_{f^{(1)}}^2(x)\right)\\
    f_{i}^{(2)}(x,n)|f^{(1)} &\iid \mathcal{N}\left(0, \sigma_{f^{(2)}}^2(x,n)\right)\\
    \sigma_{f^{(1)}}^2(x) &= \sigma_b^2 + \sigma_w^2 \frac{1}{I}\sum_{j=1}^{I}x_j^2\\
    \sigma_{f^{(2)}}^2(x,n) &= \sigma_b^2 + \sigma_w^2 \frac{1}{n}\sum_{j=1}^{n}\phi(f_j^{(1)}(x))^2.
\end{align*}
If $x'$ is another input we obtain bivariate Gaussian distributions
\begin{align*}
    (f_{i}^{(1)}(x),f_{i}^{(1)}(x')) &\iid \mathcal{N}_2\left(0, \Sigma_{f^{(1)}}(x,x')\right)\\
    (f_{i}^{(2)}(x,n),f_{i}^{(2)}(x',n))|f^{(1)} &\iid \mathcal{N}_2\left(0, \Sigma_{f^{(2)}}(x,x',n)\right),
\end{align*}
where
\begin{align*}
    \Sigma_{f^{(1)}}(x,x') &= \begin{bmatrix}\sigma_{f^{(1)}}^2(x)&c_{f^{(1)}}(x,x')\\c_{f^{(1)}}(x,x')&\sigma_{f^{(1)}}^2(x')\end{bmatrix}\\
    \Sigma_{f^{(2)}}(x,x',n) &= \begin{bmatrix}\sigma_{f^{(2)}}^2(x,n)&c_{f^{(2)}}(x,x',n)\\c_{f^{(2)}}(x,x',n)&\sigma_{f^{(2)}}^2(x',n)\end{bmatrix}\\
    c_{f^{(1)}}(x,x') &= \sigma_b^2 + \sigma_w^2 \frac{1}{I}\sum_{j=1}^{I}x_j x'_j\\
    c_{f^{(2)}}(x,x',n) &= \sigma_b^2 + \sigma_w^2 \frac{1}{n}\sum_{j=1}^{n}\phi(f_j^{(1)}(x))\phi(f_j^{(1)}(x')).
\end{align*}
Let $\overset{\mathrm{a.s.}}{\longrightarrow}$ denote the almost sure convergence. By the strong law of large numbers we know that, as $n \rightarrow +\infty$, one has
\begin{align*}
    \frac{1}{n}\sum_{j=1}^{n}\phi(f_j^{(1)}(x))^2 &\overset{\mathrm{a.s.}}{\longrightarrow} \mathbb{E}[\phi(f_1^{(1)}(x))^2]\\
    \frac{1}{n}\sum_{j=1}^{n}\phi(f_j^{(1)}(x))\phi(f_j^{(1)}(x')) &\overset{\mathrm{a.s.}}{\longrightarrow} \mathbb{E}[\phi(f_1^{(1)}(x))\phi(f_1^{(1)}(x'))],
\end{align*}
from which one can conjecture that in the limit of infinite width the stochastic processes $f_i^{(2)}(x)$ are distributed as iid (over $i$) centered GP with kernel $K(x,x')=\sigma_b^2 + \sigma_w^2\mathbb{E}[\phi(f_1^{(1)}(x))\phi(f_1^{(1)}(x'))]$. Provided that the nonlinear function $\phi$ is chosen so that $\phi(f_1^{(1)}(x))$ has finite second moment, \cite{matthews2018gaussian} made rigorous this argument and extended it to deep NNs.

A key assumption underlying the interplay between infinite wide NNs and GPs is the finiteness of the variance of the parameters' distribution at initialization. In this paper we remove the assumption of finite variance by considering iid initializations based on stable distributions, which includes Gaussian initializations as a special case. We study the infinite wide limit of fully connected feed-forward NN in the following general setting: i) the NN is deep, namely the NN is composed of multiple layers; ii) biases and scaled weights are iid according to centered symmetric stable distributions; iii) the width of network's layers goes to infinity jointly on the layers, and not sequentially on each layer; iv) the convergence in distribution is established jointly for multiple inputs, namely the convergence concerns the class of finite dimensional distributions of the NN  viewed as a stochastic process in function space. See \cite{neal1995bayesian} and \cite{der2006beyond} for early works on NNs under stable initialization.

Within our setting, we show that the infinite wide limit of the NN, under suitable scaling on the weights, is a stochastic process whose finite-dimensional distributions are multivariate stable distributions \citep{samoradnitsky2017stable}. This process is referred to as the stable process. Our result may be viewed as a generalization of the main result of \cite{matthews2018gaussian} to the context of stable distributions, as well as a generalization of results of \cite{neal1995bayesian} and \cite{der2006beyond} to the context of deep NN. Our result contributes to the theory of fully connected feed-forward deep NNs, and it paves the way to extend the research directions i) ii) and iii) that rely on Gaussian infinite wide limits. The class of stable distributions is known to be especially relevant. Indeed while the contribution of each Gaussian weight vanishes as the width grows unbounded, some of the stable weights retains significant size, thus allowing them to represent "hidden features" \citep{neal1995bayesian}.

The paper is structured as follows. \cref{sec:preliminaries} contains some preliminaries on stable distributions, whereas in \cref{sec:deep_stable_nets} we define the class of feedforward NNs considered in this work. \cref{sec:infinite_limits} contains our main result: as the width tends to infinity jointly on network's layers, the finite dimensional distributions of the NN converges to a multivariate stable distribution whose parameters are compute via a recursion over the layers. The convergence of the NN to the stable process then follows by finite-dimensional projections. In \cref{sec:related_work} we detail how our result extends previously established large width convergence results and comment on related work, whereas in \cref{sec:applications} we discuss how our result applys to the research lines highlighted in i) ii) and iii) which relies on GP limits. In \cref{sec:conclusions} we comment on future research directions. The Supplementary Material (SM) contains all the proofs (SM A,B,C), a preliminary numerical experiment on the recursion evaluation (SM D), an empirical investigation of the distribution of trained NN models' parameters (SM E). Code is available at  \url{https://github.com/stepelu/deep-stable}.

\section{Stable distributions} \label{sec:preliminaries}

Let $\text{St}(\alpha,\sigma)$ denote the symmetric centered stable distribution with stability parameter $\alpha\in(0,2]$ and scale parameter $\sigma>0$, and let $S_{\alpha,\sigma}$ be a random variable distributed as $\text{St}(\alpha,\sigma)$. That is, the characteristic function of $S_{\alpha,\sigma}\sim \text{St}(\alpha,\sigma)$ is $\varphi_{S_{\alpha,\sigma}}(t)=\mathbb{E}[\text{e}^{\textrm{i}tS_{\alpha,\sigma}}]=\text{e}^{-\sigma^{\alpha}|t|^{\alpha}}$. For any $\sigma > 0$, a $S_{\alpha,\sigma}$ random variable with $0 < \alpha < 2$ has finite absolute moments $\mathbb{E}[|S_{\alpha,\sigma}|^{\alpha -\varepsilon}]$ for any $\varepsilon > 0$, while $\mathbb{E}[|S_{\alpha,\sigma}|^{\alpha}] = + \infty$. Note that when $\alpha = 2$, we have that  $S_{2,\sigma} \sim \mathcal{N}(0, 2\sigma^2)$. The random variable $S_{2,\sigma}$ has finite absolute moments of any order. For any $a \in \mathbb{R}$ we have the scaling identity $a S_{\alpha,1} \sim \text{St}(\alpha,|a|)$.

We recall the definition of symmetric and centered multivariate stable distribution and its marginal distributions. First, let $\mathbb{S}^{k-1}$ be the unit sphere in $\mathbb{R}^{k}$. Let $\text{St}_{k}(\alpha,\Gamma)$ denote the symmetric and centered $k$-dimensional stable distribution with stability $\alpha\in(0,2]$ and scale (finite) spectral measure $\Gamma$ on $\mathbb{S}^{k-1}$, and let $\mathbf{S}_{\alpha,\Gamma}$ be a random vector of dimension $k\times 1$ distributed as $\text{St}_{k}(\alpha,\Gamma)$. The characteristic function of $\boldsymbol{S}_{\alpha,\Gamma}\sim \text{St}_k(\alpha,\Gamma)$ is
\begin{equation}\label{eq:multivariate_stable_cf}
\varphi_{\boldsymbol{S}_{\alpha,\Gamma}}(\boldsymbol{t})=\mathbb{E}[\text{e}^{\textrm{i}\boldsymbol{t}^{T}\boldsymbol{S}_{\alpha,\Gamma}}]=\exp\left\{-\int_{\mathbb{S}^{k-1}}|\boldsymbol{t}^{T}\boldsymbol{s}|^{\alpha}\Gamma(\text{d}\boldsymbol{s})\right\}.
\end{equation}
If $\boldsymbol{S}_{\alpha,\Gamma}\sim \text{St}(\alpha,\Gamma)$ then the marginal distributions of $\boldsymbol{S}_{\alpha,\Gamma}$ are described as follows. Let $\boldsymbol{1}_{r}$ denote a vector of dimension $k \times 1$ with $1$ in the $r$-the entry and $0$ elsewhere. Then, the random variable corresponding to the $r$-th element of $\boldsymbol{S}_{\alpha,\Gamma}\sim \text{St}(\alpha,\Gamma)$ can be defined as follows
\begin{equation}\label{eq:mar}
\boldsymbol{1}_{r}^{T}\boldsymbol{S}_{\alpha,\Gamma}\sim \text{St}(\alpha,\sigma(r)),
\end{equation}
where
\begin{equation}\label{eq:up_s}
\sigma(r)=\left(\int_{\mathbb{S}^{k-1}}|\boldsymbol{1}_{r}^{T}\boldsymbol{s}|^{\alpha}\Gamma(\text{d}\boldsymbol{s})\right)^{1/\alpha}.
\end{equation}
The distribution $\text{St}_{k}(\alpha,\Gamma)$ with characteristic function \cref{eq:multivariate_stable_cf} allows for marginals which are not centered nor symmetric. However in the present work all the marginals will be centered and symmetric, and the spectral measure will often be a discrete measure, i.e., $\Gamma(\cdot) = \sum_{j=1}^n \gamma_j \delta_{\boldsymbol{s}_j}(\cdot)$for $n \in \mathbb{N}$, $\boldsymbol{s}_j \in \mathbb{S}^{k-1}$ and $\gamma_j \in \mathbb{R}$. In particular, under these specific assumptions, we have
\begin{equation*}
\varphi_{\boldsymbol{S}_{\alpha,\Gamma}}(\boldsymbol{t})=\exp\left\{-\sum_{j=1}^n \gamma_j |\boldsymbol{t}^{T}\boldsymbol{s}_j|^\alpha\right\}.
\end{equation*}
See \cite{samoradnitsky2017stable} for a detailed account on $\boldsymbol{S}_{\alpha,\Gamma}\sim \text{St}(\alpha,\Gamma)$.

\section{Deep stable networks} \label{sec:deep_stable_nets}

We consider fully connected feed-forward NNs composed of $D \geq 1$ layers where each layer is of width $n \geq 1$.
Let $w_{i,j}^{(l)}$ be the weights of the $l$-th layer, and assume that they are independent and identically distributed as $\text{St}(\alpha,\sigma_{w})$, a stable distribution with stability parameter $\alpha \in (0,2]$ and scale parameter $\sigma_w > 0$. That is, the characteristic function of $
w_{i,j}^{(l)}\sim \text{St}(\alpha,\sigma_{w})$ is
\begin{equation}\label{eq:car_w}
\varphi_{w^{(l)}_{i,j}}(t)=\mathbb{E}[\text{e}^{\textrm{i}tw^{(l)}_{i,j}}]=e^{-\sigma_{w}^\alpha |t|^\alpha},
\end{equation}
for any $i\geq1$, $j\geq1$ and $l \geq 1$. Let $b_{i}^{(l)}$ denote the biases of the $l$-th hidden layer, and assume that they are independent and identically distributed as $\text{St}(\alpha,\sigma_{b})$. That is, the characteristic function of the random variable $b_{i}^{(l)}\sim \text{St}(\alpha,\sigma_{b})$ is
\begin{equation}\label{eq:car_b}
\varphi_{b^{(l)}_{i}}(t)=\mathbb{E}[\text{e}^{\textrm{i}tb^{(l)}_{i}}]=e^{-\sigma_{b}^\alpha |t|^\alpha},
\end{equation}
for any $i\geq1$ and $l \geq 1$. The random weights $w_{i,j}^{(l)}$ are independent of the biases $b_{i}^{(l)}$, for any $i\geq1$, $j\geq1$ and $l \geq 1$. That is, 
\begin{displaymath}
(w_{i,j}^{(l)}+b_{i}^{(l)})\sim \text{St}(\alpha,(\sigma_{w}^{\alpha}+\sigma_{b}^{\alpha})^{1/\alpha}).
\end{displaymath}
Let $\phi:\mathbb{R}\rightarrow\mathbb{R}$ be a nonlinearity with a finite number of discontinuities and such that it satisfies the envelope condition
\begin{equation}\label{eq:le}
|\phi(s)|\leq (a+b|s|^{\beta})^\gamma
\end{equation}
for every $s\in\mathbb{R}$, and for any parameter $a,b>0$, $\gamma<\alpha^{-1}$ and $\beta<\gamma^{-1}$. If $x\in\mathbb{R}^{I}$ is the input argument of the NN, then the NN is explicitly defined by means of
\begin{equation}\label{eq:f1}
f_{i}^{(1)}(x,n)=f_{i}^{(1)}(x)=\sum_{j=1}^{I}w_{i,j}^{(1)}x_{j}+b_{i}^{(1)},
\end{equation}
and
\begin{equation}\label{eq:fl}
f_{i}^{(l)}(x,n)=\frac{1}{n^{1/\alpha}}\sum_{j=1}^{n}w_{i,j}^{(l)}\phi(f_{j}^{(l-1)}(x,n))+b_{i}^{(l)}
\end{equation}
for $l=2,\ldots,D$ and $i=1,\ldots,n$ in \cref{eq:f1} and \cref{eq:fl}. The scaling of the weights in \cref{eq:fl} will be shown to be the correct one to obtain non-degenerate limits as $n\rightarrow+\infty$.

\section{Infinitely wide limits}\label{sec:infinite_limits}

We show that, as the width of the NN tends to infinity jointly on network's layers, the finite dimensional distributions of the network converge to a multivariate stable distribution whose parameters are compute via a suitable recursion over the network layers. Then, by combining this limiting result with standard arguments on finite-dimensional projections we obtain the large $n$ limit of the stochastic process $(f_i^{(l)}(x^{(1)},n),\dots,f_i^{(l)}(x^{(k)},n))_{i \geq 1}$ where $x^{(1)},\dots,x^{(k)}$ are the inputs to the NN. In particular, let $\overset{w} {\longrightarrow}$ denote the weak convergence. Then, we show that as $n\rightarrow+\infty$,
\begin{equation}\label{eq:main_informal}
(f_i^{(l)}(x^{(1)},n),\dots,f_i^{(l)}(x^{(k)},n))_{i \geq 1} \overset{w} {\longrightarrow} \bigotimes_{i\geq1}\text{St}_{k}(\alpha,\Gamma(l))
\end{equation}
where $\bigotimes$ is the product measure. From now on $k$ is the number of inputs, which is equal to the dimensionality of the finite dimensional distributions of interest for the stochastic processes $f_i^{(l)}$.Thorough the rest of the paper we assume that the assumptions introduced in \cref{sec:deep_stable_nets} hold true. Hereafter we present a sketch of the proofs of our main result for a fixed index $i$ and input $x$, and we defer to the SM for the complete proofs of our main results.

We start with a technical remark: in \cref{eq:f1}-\cref{eq:fl} the stochastic processes $f_i^{(l)}(x,n)$ are only defined for $i=1,\dots,n$,  while the limiting measure in \cref{eq:main_informal} is the product measure on $i \geq 1$. This fact does not determine problems, as for each $\mathcal{L} \subset \mathbb{N}$ there is a $n$ large enough such that for each $i \in \mathcal{L}$ the processes $f_i^{(l)}(x,n)$ are defined. In any case, the simplest solution consists in extending $f_i^{(l)}(x,n)$ from $i=1,\dots,n$ to $i \geq 1$ in \cref{eq:f1}-\cref{eq:fl}, and we will make this assumption in all the proofs.

\subsection{Large width asymptotics: \texorpdfstring{$k=1$}{k=1}}

We characterize the limiting distribution of $f_i^{(l)}(x,n)$ as $n \rightarrow \infty$ for a fixed $i$ and input $x$. We show that, as $n\rightarrow+\infty$,
\begin{equation}\label{eq:convergence_1}
{f}_{i}^{(l)}(x,n)\overset{w}{\longrightarrow} \text{St}(\alpha,\sigma(l)),
\end{equation}
where the parameter $\sigma(l)$ is computed through the recursion:
\begin{align*}
    \sigma(1) &= \big(\sigma_b^\alpha + \sigma_w^\alpha \sum_{j=1}^I |x_j|^\alpha\big)^{1/\alpha}\\
    \sigma(l) &= \big(\sigma_b^\alpha + \sigma_w^\alpha \mathbb{E}_{f \sim q^{(l-1)}}[|\phi(f)|^\alpha]\big)^{1/\alpha}
\end{align*}
and $q(l) = \text{St}(\alpha,\sigma(l))$ for each $l \geq 1$. The generalization of this result to $k \geq 1$ inputs is given in \cref{sec:asympt_k_brief}.
\begin{proof}[Proof of \cref{eq:convergence_1}] The proof exploits exchangeability of the sequence $({f}_i^{(l)}(n,x))_{i\geq1}$, an induction argument  on the layer's index $l$ for the directing (random measure) of $({f}_i^{(l)}(n,x))_{i\geq1}$, and some technical lemmas that are proved in SM. Recall that the input is a real-valued vector of dimension $I$.
By means of \cref{eq:car_w} and \cref{eq:car_b}, for $i \geq 1$:
\begin{align*}
&\varphi_{f_{i}^{(1)}(x)}(t)\\
&=\mathbb{E}[e^{\textrm{i}tf_{i}^{(1)}(x)}]\\
&=\mathbb{E}\left[\exp\left\{\textrm{i}t\left[\sum_{j=1}^{I}w_{i,j}^{(1)}x_{j}+b_{i}^{(1)}\right]\right\}\right]\\
&=\exp\left\{-(\sigma_{w}^{\alpha}\sum_{j=1}^{I}|x_{j}|^{\alpha}+\sigma_{b}^{\alpha})|t|^{\alpha}\right\},
\end{align*}
i.e.
\begin{equation*}
f_{i}^{(1)}(x)\overset{\text{d}}{=}S_{\alpha,\left(\sigma_{w}^{\alpha}\sum_{j=1}^{I}|x_{j}|^{\alpha}+\sigma_{b}^{\alpha}\right)^{1/\alpha}};
\end{equation*}
and for $l=2,\dots,D$
\begin{align*}
&\varphi_{f_{i}^{(l)}(x,n)\,|\,\{f_{j}^{(l-1)}(x,n)\}_{j=1,\ldots,n}}(t)\\
&=\mathbb{E}[e^{\textrm{i}tf_{i}^{(l)}(x,n)}\,|\,\{f_{j}^{(l-1)}(x,n)\}_{j=1,\ldots,n}]\\
&=\mathbb{E}\Bigg[\exp\Bigg\{\textrm{i}t\Bigg[\frac{1}{n^{1/\alpha}}\sum_{j=1}^{n}w_{i,j}^{(l)}\phi(f_{j}^{(l-1)}(x,n))+b_{i}^{(l)}\Bigg]\Bigg\}\\
&\qquad \mathrel{\Bigg|} \{f_{j}^{(l-1)}(x,n)\}_{j=1,\ldots,n}\Bigg]\\
&=\exp\left\{-(\frac{\sigma_{w}^{\alpha}}{n}\sum_{j=1}^{n}|\phi(f_{j}^{(l-1)}(x,n))|^{\alpha}+\sigma_{b}^{\alpha})|t|^{\alpha}\right\},
\end{align*}
i.e.,
\begin{align*}
&f_{i}^{(l)}(x,n)\,|\,\{f_{j}^{(l-1)}(x,n)\}_{j=1,\ldots,n}\\
&\notag\overset{\text{d}}{=} S_{\alpha,\left(\frac{\sigma_{w}^{\alpha}}{n}\sum_{j=1}^{n}|\phi(f_{j}^{(l-1)}(x,n))|^{\alpha}+\sigma_{b}^{\alpha}\right)^{1/\alpha}}.
\end{align*}
It comes from \eqref{eq:fl} that, for every fixed $l$ and for every fixed $n$ the sequence $({f}_i^{(l)}(n,x))_{i\geq1}$ is exchangeable. In particular, let $p_{n}^{(l)}$ denote the directing (random) probability measure of the exchangeable sequence $({f}_i^{(l)}(n,x))_{i\geq1}$. That is, by de Finetti representation theorem, conditionally to $p_{n}^{(l)}$ the ${f}_i^{(l)}(n,x)$'s are iid as $p_{n}^{(l)}$. Now, consider the induction hypothesis that $p_{n}^{(l-1)}\stackrel{w}{\longrightarrow}q^{(l-1)}$ as $n\rightarrow+\infty$, with $q^{(l-1)}$ be $\text{St}(\alpha,\sigma(l-1))$, and the parameter $\sigma(l-1)$ will be specified. Therefore,
\begin{align}\label{eq_princi}
&\notag\mathbb{E}[\text{e}^{\textrm{i}t{f}_i^{(l)}(x,n)}]\\
&\notag=\mathbb{E}\left[\exp\left\{-|t|^\alpha\left(\frac{\sigma_{w}^\alpha}{n}\sum_{j=1}^n|\phi({f}_j^{(l-1)}(x,n))|^\alpha+\sigma^{\alpha}_{b}\right)\right\}\right]\\
&\notag=e^{-|t|^{\alpha}\sigma_{b}^{\alpha}}\mathbb{E}\left[\exp\left\{-|t|^\alpha\frac{\sigma_{w}^\alpha}{n}\sum_{j=1}^n|\phi({f}_j^{(l-1)}(x,n))|^\alpha\right\}\right]\\
&=e^{-|t|^{\alpha}\sigma_{b}^{\alpha}}\mathbb{E}\left[\left(\int \exp\left\{-|t|^\alpha\frac{\sigma_{w}^\alpha}{n}|\phi({f})|^\alpha\right\}p_{n}^{(l-1)}(\text{d}{f})\right)^n\right].
\end{align}
where the first equality comes from plugging in the definition of $f_i^{(l)}(x,n)$, rewriting $\mathbb{E}[\exp(\sum_{j=1}^n\cdots)]$ as $\mathbb{E}[\prod_{j=1}^n\exp(\cdots)] = \prod_{j=1}^n\mathbb{E}[\exp(\cdots)]$ due to independence, computing the characteristic function for each term, and re-arranging. therein, since $(f_{I}^{(l-1)}(n,x))_{i\geq1}$ is exchangeable there exists (de Finetti theorem) a random probability measure $p_{n}^{(l-1)}$ such that conditionally to $p_{n}^{(l-1)}$ the $f_{I}^{(l-1)}(n,x)$ are iid as $p_{n}^{(l-1)}$ which explains \cref{eq_princi}.

Now, let $\overset{p} {\longrightarrow}$ denote the convergence in probability. The following technical lemmas (\cref{sec:asympt_1}):
\begin{itemize}
	\item[L1)] for each $l \geq 2$ $\text{Pr}[p_{n}^{(l-1)}\in I]=1$, with $I=\{p:\int |\phi({f})|^\alpha p(\text{d}{f})<+\infty\}$;
	\item[L2)] $\int |\phi({f})|^\alpha p_{n}^{(l-1)}(\text{d}{f})\stackrel{p}{\longrightarrow} \int |\phi({f})|^\alpha q^{(l-1)}(\text{d}{f})$, as $n\rightarrow +\infty$;
	\item[L3)]  $\int |\phi({f})|^\alpha [1-e^{-|t|^\alpha\frac{\sigma^{\alpha}_{w}}{n} |\phi({f})|^\alpha}]p_{n}^{(l-1)}(\text{d}{f})\stackrel{p}{\longrightarrow} 0$, as $n\rightarrow +\infty$.
\end{itemize}
together with Lagrange theorem, are the main ingredients for proving \cref{eq:convergence_1} by studying the large $n$ asymptotic behavior of \cref{eq_princi}. By combining \eqref{eq_princi} with lemma L1,
\begin{align*}
&\mathbb{E}[\text{e}^{\textrm{i}t{f}_i^{(l)}(x,n)}] =e^{-|t|^{\alpha}\sigma_{b}^{\alpha}} \mathbb{E}\Bigg[\mathbbm{1}_{\{(p_{n}^{(l-1)}\in I)\}}
\\&\times\Bigg(\int \exp\Bigg\{-|t|^\alpha\frac{\sigma_{w}^\alpha}{n}|\phi({f})|^\alpha\Bigg\}p_{n}^{(l-1)}(\text{d}{f})\Bigg)^n\Bigg].
\end{align*}
By means of Lagrange theorem, there exists $\theta_{n}\in[0,1]$ such that
\begin{align*}
&\exp\left\{-|t|^\alpha\frac{\sigma_{w}^\alpha}{n}|\phi({f})|^\alpha\right\}\\
&\quad=1-|t|^\alpha\frac{\sigma_{w}^\alpha}{n}|\phi({f})|^\alpha\\
&\quad\quad+|t|^\alpha\frac{\sigma_{w}^\alpha}{n}|\phi({f})|^{\alpha}\left(1-\exp\left\{-\theta_{n}|t|^\alpha\frac{\sigma_{w}^\alpha}{n}|\phi({f})|^\alpha\right\}\right).
\end{align*}
Now, since
\begin{align*}
0&\leq \int|\phi({f})|^{\alpha}[1-e^{-\theta_{n}|t|^\alpha\frac{\sigma^{\alpha}_{w}}{n} |\phi({f})|^\alpha}]p_{n}^{(l-1)}(\text{d}{f})\\
&\leq \int|\phi({f})|^{\alpha}[1-e^{-|t|^\alpha\frac{\sigma^{\alpha}_{w}}{n} |\phi({f})|^\alpha}]p_{n}^{(l-1)}(\text{d}{f}),
\end{align*}
\begin{align*}
&\mathbb{E}[\text{e}^{\textrm{i}t{f}_i^{(l)}(x,n)}] \leq e^{-|t|^{\alpha}\sigma_{b}^{\alpha}}\mathbb{E}\Bigg[\mathbbm{1}_{\{(p_{n}^{(l-1)}\in I)\}}\\
&\times\Bigg(1-|t|^{\alpha}\frac{\sigma^{\alpha}_{w}}{n}\int|\phi({f})|^{\alpha}p_{n}^{(l-1)}(\text{d}{f})\Bigg.\Bigg.\\
&+\Bigg.\Bigg.|t|^{\alpha}\frac{\sigma^{\alpha}_{w}}{n}\int|\phi({f})|^{\alpha}[1-e^{-|t|^\alpha\frac{\sigma^{\alpha}_{w}}{n} |\phi({f})|^\alpha}]p_{n}^{(l-1)}(\text{d}{f})\Bigg)^{n}\Bigg].
\end{align*}
Finally, recall the fundamental limit $\text{e}^{x}=\lim_{n\rightarrow+\infty}(1+x/n)^{n}$. This, combined with L2 and L3 leads to 
\begin{align*}
&\mathbb{E}[\text{e}^{\textrm{i}t{f}_i^{(l)}(x,n)}]\rightarrow e^{-|t|^\alpha[\sigma^{\alpha}_{b}+\sigma^\alpha_{w} \int |\phi({f})|^\alpha q^{(l-1)}(\text{d}{f})]},
\end{align*}
as $n\rightarrow+\infty$. That is, we proved that the large $n$ limiting distribution of ${f}_i^{(l)}(x,n)$ is $\text{St}(\alpha,\sigma(l))$, where we set 
\begin{equation*}
\sigma(l)=\left(\sigma^{\alpha}_{b}+\sigma^\alpha_{w} \int |\phi({f})|^\alpha q^{(l-1)}(\text{d}{f})\right)^{1/\alpha}
\end{equation*}
\end{proof}

\subsection{Large width asymptotics: \texorpdfstring{$k \geq 1$}{k>=1}}\label{sec:asympt_k_brief}

We establish the convergence in distribution of $(f_i^{(l)}(x^{(1)},n),\dots,f_i^{(l)}(x^{(k)},n))$ as $n \rightarrow +\infty$ for a fixed $i$ and $k$ inputs $x^{(1)},\dots,x^{(k)}$. This result, combined with standard arguments on finite-dimensional projections, then establishes the convergence of the NN to the stable process. Precisely, we show that, as $n\rightarrow+\infty$, one has
\begin{equation}\label{eq:convergence_k}
(f_i^{(l)}(x^{(1)},n),\dots,f_i^{(l)}(x^{(k)},n))\overset{w}{\longrightarrow}\text{St}_{k}(\alpha,\Gamma(l)),
\end{equation}
where the spectral measure $\Gamma(l)$ is computed through the recursion:
\begin{align}
    \Gamma(1) &= \sigma_b^\alpha ||1||^\alpha \delta_{\frac{\boldsymbol{1}}{||\boldsymbol{1}||}} + \sigma_w^\alpha \sum_{j=1}^I ||\boldsymbol{x}_j||^\alpha \delta_{\frac{\boldsymbol{x}_j}{||\boldsymbol{x}_j||}}\label{eq:recursion_1}\\
    \Gamma(l) &= \sigma_b^\alpha ||1||^\alpha \delta_{\frac{\boldsymbol{1}}{||\boldsymbol{1}||}} + \sigma_w^\alpha \mathbb{E}_{f \sim q^{(l-1)}}[||\phi(f)||^\alpha\delta_{\frac{\phi(f)}{||\phi(f)||}}]\label{eq:recursion_l}
\end{align}
and $q(l) = \text{St}_{k}(\alpha,\Gamma(l))$ for each $l \geq 1$, where $\boldsymbol{x}_j = [x^{(1)}_j,\dots,x^{(k)}_j] \in \mathbb{R}^k$. Here (and in all the expressions involving the function $\delta$ ) we make use of the notational assumption that if $\lambda = 0$ in $\lambda \delta_{\bullet}$, then $\lambda \delta_{\bullet} = 0$. This assumption allows us to avoid making the notation more cumbersome than necessary to explicitly exclude the case of $\phi(f) = 0$, for which $\phi(f)/||\phi(f)||$ is undefined. We omit the sketch of the proof of \cref{eq:convergence_k}, as it is a step-by-step parallel of the proof of \cref{eq:convergence_1} with the added complexities due to the multivariate stable distributions. The reader can refer to the SM for the full proof.

\subsection{Finite-dimensional projections}
In \cref{sec:asympt_k_brief} we obtained the convergence in law of $f_i^{(l)}(x^{(1)},n),\dots,f_i^{(l)}(x^{(k)},n)$ for $k$ inputs and a generic $i$ to a multivariate Stable distribution. Let refer to this random vector as $f_i(x)$.  Now, we derive the limiting behavior in law of $f_i(x)$ jointly over all $i=1,\dots$ (again for a given $k$-dimensional input). It is enough to study the convergence of $f_1(x),\dots,f_n(x)$ for a generic $n \geq 1$. That is, it is enough to establish the convergence of the finite dimensional distributions (over $i$: we consider here $f_i(x)$ as random sequence over $i$). See \cite{billingsley1999convergence} for details. 

To establish the convergence of the finite dimensional distributions (over $i$) it then suffices to establish the convergence of linear combinations. More precisely, let $\boldsymbol{X} = [x^{(1)},\dots,x^{(k)}] \in \mathbb{R}^{I \times k}$. We show that, as $n\rightarrow+\infty$,
\begin{equation*}
({f}_{i}^{(l)}(\boldsymbol{X},n))_{i\geq1}\overset{w}{\longrightarrow}\bigotimes_{i\geq1}\text{St}_{k}(\alpha,\Gamma(l)),
\end{equation*}
by proving the large $n$ asymptotic behavior of any finite linear combination of the ${f}_{i}^{(l)}(\boldsymbol{X},n)$'s, for $i\in\mathcal{L}\subset\mathbb{N}$.  Following the notation of \cite{matthews2018gaussian}, let 
\begin{equation*}
T^{(l)}(\mathcal{L},p,\boldsymbol{X},n)=\sum_{i\in \mathcal{L}}p_{i}[{f}_{i}^{(l)}(\boldsymbol{X},n)-b_{i}^{(l)}\boldsymbol{1}].
\end{equation*}
Then, we write
\begin{align*}
&T^{(l)}(\mathcal{L},p,\boldsymbol{X},n)\\
&=\sum_{i\in \mathcal{L}}p_{i}\left[\frac{1}{n^{1/\alpha}}\sum_{j=1}^{n}w_{i,j}^{(l)}(\phi\circ {f}_{j}^{(l-1)}(\boldsymbol{X},n))\right]\\
&=\frac{1}{n^{1/\alpha}}\sum_{j=1}^{n}\gamma_{j}^{(l)}(\mathcal{L},p,\boldsymbol{X},n),
\end{align*}
where
\begin{equation*}
\gamma_{j}^{(l)}(\mathcal{L},p,\boldsymbol{X},n)=\sum_{i\in \mathcal{L}}p_{i}w_{i,j}^{(l)}(\phi\circ{f}_{j}^{(l-1)}(\boldsymbol{X},n)).
\end{equation*}
Then,
\begin{align*}
&\varphi_{T^{(l)}(\mathcal{L},p,\mathbf{X},n)\,|\,\{{f}_{j}^{(l-1)}(\boldsymbol{X},n)\}_{j=1,\ldots,n}}(\boldsymbol{t})\\
&=\mathbb{E}[e^{\textrm{i}\boldsymbol{t}^{T}T^{(l)}(\mathcal{L},p,\boldsymbol{X},n)}\,|\,\{{f}_{j}^{(l-1)}(\boldsymbol{X},n)\}_{j=1,\ldots,n}]\\
&=\prod_{j=1}^{n}\prod_{i\in\mathcal{L}}\text{e}^{-\frac{p^{\alpha}_{i}\sigma_{w}^{\alpha}}{n}|\boldsymbol{t}^{T}(\phi\circ f_{j}^{(l-1)}(\boldsymbol{X},n))|^{\alpha}}
\end{align*}
That is, 
\begin{align*}
&T^{(l)}(\mathcal{L},p,\boldsymbol{X},n)\,|\,\{{f}_{j}^{(l-1)}(\boldsymbol{X},n)\}_{j=1,\ldots,n}\overset{\text{d}}{=}\boldsymbol{S}_{\alpha,\Gamma^{(l)}}
\end{align*}
where
\begin{equation*}
\Gamma^{(l)}=\frac{1}{n}\sum_{j=1}^{n}\sum_{i\in\mathcal{L}}||p_{i}\sigma_{w}(\phi\circ f_{j}^{(l-1)}(\boldsymbol{X},n))||^{\alpha}\delta_{\frac{\phi\circ f_{j}^{(l-1)}(\boldsymbol{X},n)}{||\phi\circ f_{j}^{(l-1)}(\boldsymbol{X},n)||}}
\end{equation*}
Then, along lines similar to the proof of the large $n$ asymptotics for the $i$-th coordinate, we have the following
\begin{align*}
&\mathbb{E}[\text{e}^{\textrm{i}\boldsymbol{t}^{T}T^{(l)}(\mathcal{L},p,\boldsymbol{X},n)}] \rightarrow \exp\Bigg\{-\int\int_{\mathbb{S}^{k-1}}|\boldsymbol{t}^{T}\boldsymbol{s}|^{\alpha}\\
& \times \Bigg(\sum_{i\in\mathcal{L}}||p_{i}\sigma_{w}(\phi\circ{f})||^{\alpha}\delta_{\frac{\phi\circ{f}}{||\phi\circ{f}||}}\Bigg)(\text{d}\boldsymbol{s})q^{(l-1)}(\text{d}{f})\Bigg\}
\end{align*}
as $n\rightarrow+\infty$. This complete the proof of the limiting behaviour \eqref{eq:main_informal}.

\section{Related work}\label{sec:related_work}

For the classical case of Gaussian weights and biases, and more in general for finite-variance iid distributions, the seminal work is that of \cite{neal1995bayesian}. Here the author establishes, among other notable contributions, the connection between infinitely wide shallow (1 hidden layer) NNs and centered GPs. We reviewed the essence of it in \cref{sec:intro}.

This result is extended in \cite{lee2018deep} to deep NNs where the width $n(l)$ of each layer $l$ goes to infinity sequentially, starting from lowest layer, i.e. $n(1)$ to $n(D)$. The sequential nature of the limits reduces the task to a sequential application of the approach of \cite{neal1995bayesian}. The computation of the GP kernel for each layer $l$ involves a recursion, and the authors propose a numerical method to approximate the integral involved in each step of the recursion. The case where each $n(l)$ goes to infinity jointly, i.e. $n(l)=n$, is considered in \cite{g.2018gaussian} under more restrictive hypothesis, which are relaxed in \cite{matthews2018gaussian}. While this setting is most representative of a sequence of increasingly wide networks, the theoretical analysis is considerably more complicated as it does not reduce to a sequential application of the classical multivariate central limit theorem.

Going beyond finite-variance weight and bias distributions, \cite{neal1995bayesian} also introduced preliminary results for infinitely wide shallow NNs when weights and biases follow centered symmetric stable distributions. These results are refined in \cite{der2006beyond} which establishes the convergence to a stable process, again in the setting of shallow NNs.

The present paper can be considered a generalization of the work of \cite{matthews2018gaussian} to the context of weights and biases distributed according to centered and symmetric stable distributions. Our proof follows different arguments from the proof of  \cite{matthews2018gaussian}, and in particular it does not rely on the central limit theorem for exchangeable sequences (Blum et al., 1958). Hence, since the Gaussian distribution is a special case of the stable distribution, our proof provides an alternative and self-contained proof to the result of \cite{matthews2018gaussian}. It should be noted that our envelope condition \cref{eq:le} is more restrictive than the linear envelope condition of  \cite{matthews2018gaussian}. For the classical Gaussian setting the conditions on the activation function have been weakened in the work of \cite{1yang2019scaling}.

Finally, there has been recent interest in using heavy-tailed distributions for gradient noise \citep{simsekli2019tail} and for trained parameter distributions \citep{martin2019traditional}. In particular, \cite{martin2019traditional} includes an empirical analysis of the parameters of pre-trained convolutional architectures (which we also investigate in SM E) supportive of heavy-tailed distributions. Results of this kind are compatible with the conjecture that stochastic processes arising from NNs whose parameters are heavy-tailed might be closer representations of their finite, high-performing, counterparts.

\section{Future applications}\label{sec:applications}

\subsection{Bayesian inference}\label{sec:bayesian_inference}

Infinitely wide NNs with centered iid Gaussian initializations, and more in general finite variance centered iid initializations, gives rise to iid centered GPs at every layer $l$. Let us assume that weights and biases are distributed as in \cref{sec:intro}, and let us assume $L$ layers ($L-1$ hidden layers). Each centered GPs is characterized by its covariance kernel function. Let us denote by $f^{(l)}$ such GPs for the layer $2 \leq l \leq L$. Over two inputs $x$ and $x'$ the distribution of $(f^{(l)}(x),f^{(l)}(x'))$ is characterized by the variances $q^{(l)}_x = \mathbb{V}[f^{(l)}(x)]$, $ q^{(l)}_{x'} = \mathbb{V}[f^{(l)}(x')]$ and by the covariance $c^{(l)}_{x,x'} = \mathbb{C}[f^{(l)}(x),f^{(l)}(x')]$. These quantities satisfy
\begin{align}
    q^{(l)}_x &= \sigma_b^2 + \sigma_w^2 \mathbb{E}\Bigg[\phi\Bigg(\sqrt{q^{(l-1)}_x}z\Bigg)^2\Bigg]\label{eq:ip_q}\\
    c^{(l)}_{x,x'} &= \sigma_b^2 + \sigma_w^2\mathbb{E}\Bigg[\phi\Bigg(\sqrt{q^{(l-1)}_x}z\Bigg)\label{eq:ip_c}\\
    &\notag\times\phi\Bigg(\sqrt{q^{(l-1)}_{x'}}\Big(\rho^{(l-1)}_{x,x'}z + \sqrt{1 - (\rho^{(l-1)}_{x,x'})^2}z'\Big)\Bigg)\Bigg]
\end{align}
where $z$ and $z'$ are independent standard Gaussian distributions $\mathcal{N}(0,1)$,
\begin{equation}
    \rho^{(l)} = \frac{c^{(l)}_{x,x'}}{\sqrt{q^{(l)}_x q^{(l)}_{x'}}}\label{eq:ip_rho}
\end{equation}
with  initial conditions $q^{(1)}_x = \sigma_b^2 + \sigma_w^2 \norm{x}^2$ and $c^{(1)}_{x,x'} = \sigma_b^2 + \sigma_w^2 \dprod{x}{x'}$.

To perform prediction via $\mathbb{E}[f^{(L)}(x^{*})|x^{*},\mathcal{D}]$, it is necessary to compute these recursions for all ordered pairs of data points $x,x'$ in the training dataset $\mathcal{D}$, and for all pairs $x^{*},x$ with $x\in\mathcal{D}$. \cite{lee2018deep} proposes an efficient quadrature solution that keeps the computational requirements manageable for an arbitrary activation $\phi$.

In our setting, the corresponding recursion is defined by \cref{eq:recursion_1}-\cref{eq:recursion_l}, which is a more computationally challenging problem with respect to the Gaussian setting. A sketch of a potential approach is as follows. Over the training data points and test points, $f^{(1)} \sim \text{St}_k(\alpha,\Gamma(1))$ where $k$ is equal to the size of training and test datasets combined. As $\Gamma(1)$ is a discrete measure exact simulations algorithms are available with a computational cost of $\mathcal{O}(I)$ per sample \citep{nolan2008overview}. We can thus generate $M$ samples $\widetilde{f}^{(1)}_j$, $j=1,\dots,M$, in $\mathcal{O}(IM)$, and use these to approximate $f^{(2)} \sim \text{St}_k(\alpha,\Gamma(2))$ with $\text{St}_k(\alpha,\widetilde{\Gamma}(2))$ with $\widetilde{\Gamma}(2)$ being
\begin{equation*}
    \widetilde{\Gamma}(2) = \sigma_b^\alpha ||1||^\alpha \delta_{\frac{\boldsymbol{1}}{||\boldsymbol{1}||}} + \sigma_w^\alpha \sum_{j=1}^M ||\phi(\widetilde{f}^{(1)}_j)||^\alpha\delta_{\frac{\phi(\widetilde{f}^{(1)}_j)}{||\phi(\widetilde{f}^{(1)}_j)||}}
\end{equation*}
We can repeat this procedure by generating (approximate) random samples $\widetilde{f}^{(2)}_j$, with a cost of $\mathcal{O}(M^2)$, that in turn are used to approximate $\Gamma(3)$ and so on. In this procedure the errors can accumulate across the layers, as in \cite{lee2018deep}. This may be ameliorated by using quasi random number generators of \cite{joe2008notes}, as the sampling algorithms for multivariate stable distributions  \citep{weron1996chambers,weron2010correction,nolan2008overview} are all implemented as transformations of uniform distributions. The use of QRNG effectively defines a quadrature scheme for the integration problem. We report in the SM preliminary results regarding the numerical approximation of the recursion defined by \cref{eq:recursion_1}-\cref{eq:recursion_l}.

This leaves us with the problem of computing a statistic of $f^{(L)}(x^{*})|(x^{*},\mathcal{D})$ or sampling from it, to perform prediction. Again, it could be beneficial to leverage on the discreteness of $\widetilde{\Gamma}(L)$. For example, these multivariate stable random variables can be expressed as suitable linear transformations of independent stable random variables \citep{samoradnitsky2017stable}, and results expressing stable variables as mixtures of Gaussian variables are available in \cite{samoradnitsky2017stable}.

\subsection{Neural tangent kernel}

In \cref{sec:bayesian_inference} we reviewed how the connection with GPs makes it possible to perform Bayesian inference directly on the limiting process. This corresponds to a "weakly-trained" regime of NNs, in the sense that the point (mean) predictions are equivalent to assuming an $l_2$ loss function, and fitting only a terminal linear layer to the training data, i.e. performing a kernel regression \citep{arora2019exact}. The works of \cite{jacot2018neural}, \cite{lee2019wide} and \cite{arora2019exact} consider "fully-trained" NNs with $l_2$ loss and continuous-time gradient descent. Under Gaussian initialization assumptions it is shown that as the width of the NN goes to infinity, the point predictions corresponding by such fully trained networks are given again by a kernel regression but with respect to a different kernel, the neural tangent kernel.

In the derivation of the neural tangent kernel, one important point is that the gradients are not computed with respect to the standard model parameters, i.e. the the weights and biases entering the affine transforms. Instead they are "reparametrized gradients" which are computed with respect to parameters initialized as $\mathcal{N}(0,1)$, with any scaling (standard deviation) defined by parameter multiplication. It would thus be interesting to study whether a corresponding neural tangent kernel can be defined for the case of stable distributions with $0 < \alpha < 2$, and whether the parametrization of \cref{eq:f1}-\cref{eq:fl} is the appropriate one to do so.

\subsection{Information propagation}

The recursions \cref{eq:ip_q}-\cref{eq:ip_c} define the evolution over depth of the distribution of $f^{(l)}$ for two points $x,x'$ when weights and biases are distributed as in \cref{sec:intro}. The information propagation framework studies the behavior of $q_x^{(l)}$ and $\rho_{x,x'}^{(l)}$ as $l \rightarrow +\infty$. It is shown in \cite{poole2016exponential} and \cite{schoenholz2017deep} that the $(\sigma_w,\sigma_b)$ positive quadrant is divided in two regions: a stable phase where $\rho_{x,x'}^{(l)} \rightarrow 1$ and a chaotic phase where $\rho_{x,x'}^{(l)}$ converges to a random variable (in the $\phi=\tanh$ case, in other cases the limiting processes may fail to exist). Thus in the stable phase $f^{(l)}$ is eventually perfectly correlated over inputs (and in most cases perfectly constant), while in the chaotic phase it is almost everywhere discontinuous. The work of \cite{hayou2019impact} formalizes these results and investigates the case where $(\sigma_w,\sigma_b)$ is on the curve separating the stable from the chaotic phase, i.e. the edge of chaos. Here it is shown that the behavior is qualitatively similar to that of the stable case, but with a lower rate of convergence with respect to depth. Thus in all cases the distribution of $f^{(l)}$ eventually collapse to degenerate and inexpressive distributions as depth increases.

In this context it would be interesting to study what is the impact of the use of stable distributions. All results mentioned above holds for the Gaussian case, which corresponds to $\alpha=2$. Thus this further analysis would study the case $0 < \alpha < 2$, resulting in a triplet $(\sigma_w,\sigma_b,\alpha)$. Even though it seems hard to escape the course of depth under iid initializations, it might be that the use of stable distributions, with their not-uniformly-vanishing relevance at unit level \citep{neal1995bayesian}, might slow down the rate of convergence to the limiting regime.

\section{Conclusions}\label{sec:conclusions}

Within the setting of fully connected feed-forward deep NNs with weights and biases iid as centered and symmetric stable distributions, we proved that the infinite wide limit of the NN, under suitable scaling on the weights, is a stable process. This result contributes to the theory of fully connected feed-forward deep NNs, generalizing the work of \cite{matthews2018gaussian}. We presented an extensive discussion on how our result can be used to extend recent lines of research which relies on GP  limits. 

On the theoretical side further developments of our work are possible. Firstly, \cite{matthews2018gaussian} performs an empirical analysis of the rates of convergence to the limiting process as function of depth with the respect to the MMD discrepancy \citep{gretton2012kernel}. Having proved the convergence of the finite dimensional distributions to multivariate stable distributions, the next step would be to establish the rate of convergence with respect to a metric of choice as function of the stability index $\alpha$ and depth $l$. Secondly, all the established convergence results (this paper included) concern the convergence of the finite dimensional distributions of the NN layers. For the countable case, which is the case of the components $i \geq 1$ in each layer, this is equivalent to the convergence in distribution of the whole process (over all the $i$) with respect to the product topology. However, the input space being $R^I$ it is not countable. Hence, for a given $i$, the convergence of the finite dimensional distributions (i.e. over a finite collection of inputs) is not enough to establish the convergence in distribution of the stochastic process seen as a random function on the input (with respect to an appropriate metric). This is also the case for results concerning the convergence to GPs. It would thus be worthwhile to complete this theoretical line of research by establishing such result for any $0 < \alpha \leq 2$. As a side result, doing so is likely to provide estimates on the smoothness proprieties of the limiting stochastic processes.

\section{Acknowledgements}\label{sec:acknowledgements}

We wish to thank the three anonymous reviewers and the meta reviewer for their valuable feedback. Stefano Favaro received funding from the European Research Council (ERC) under the European Union's Horizon 2020 research and innovation programme under grant agreement No 817257. Stefano Favaro gratefully acknowledge the financial support from the Italian Ministry of Education, University and Research (MIUR), ``Dipartimenti di Eccellenza" grant 2018-2022.

\bibliography{ref}
\bibliographystyle{apalike}

\clearpage
\appendix
\onecolumn

\section{Large width asymptotics: \texorpdfstring{$k=1$}{k=1}} \label{sec:asympt_1}
We first consider the case with of a single input being a real-valued vector of dimension $I$.
By means of \cref{eq:car_w} and \cref{eq:car_b} we can write for $i \geq 1$ and $l=2,\dots,D$:
\begin{itemize}
\item[i)]
\begin{align*}
\varphi_{f_{i}^{(1)}(x)}(t)&=\mathbb{E}[e^{\textrm{i}tf_{i}^{(1)}(x)}]\\
&=\mathbb{E}\left[\exp\left\{\textrm{i}t\left[\sum_{j=1}^{I}w_{i,j}^{(1)}x_{j}+b_{i}^{(1)}\right]\right\}\right]\\
&=\mathbb{E}[\exp\{\textrm{i}tb_{i}^{(1)}\}]\prod_{j=1}^{I}\mathbb{E}[\exp\{\textrm{i}tw_{i,j}^{(1)}x_{j}\}]\\
&=\text{e}^{-\sigma_{b}^{\alpha}|t|^{\alpha}}\prod_{j=1}^{I}\text{e}^{-\sigma_{w}^{\alpha}|tx_{j}|^{\alpha}}\\
&=\exp\left\{-(\sigma_{w}^{\alpha}\sum_{j=1}^{I}|x_{j}|^{\alpha}+\sigma_{b}^{\alpha})|t|^{\alpha}\right\},
\end{align*}
i.e.,
\begin{equation*}
f_{i}^{(1)}(x)\overset{\text{d}}{=}S_{\alpha,\left(\sigma_{w}^{\alpha}\sum_{j=1}^{I}|x_{j}|^{\alpha}+\sigma_{b}^{\alpha}\right)^{1/\alpha}};
\end{equation*}
\item[ii)]
\begin{align*}
&\varphi_{f_{i}^{(l)}(x,n)\,|\,\{f_{j}^{(l-1)}(x,n)\}_{j=1,\ldots,n}}(t)\\
&\quad=\mathbb{E}[e^{\textrm{i}tf_{i}^{(l)}(x,n)}\,|\,\{f_{j}^{(l-1)}(x,n)\}_{j=1,\ldots,n}]\\
&\quad=\mathbb{E}\left[\exp\left\{\textrm{i}t\left[\frac{1}{n^{1/\alpha}}\sum_{j=1}^{n}w_{i,j}^{(l)}\phi(f_{j}^{(l-1)}(x,n))+b_{i}^{(l)}\right]\right\}\,|\,\{f_{j}^{(l-1)}(x,n)\}_{j=1,\ldots,n}\right]\\
&\quad=\mathbb{E}[\exp\{\textrm{i}tb_{i}^{(l)}\}]\prod_{j=1}^{n}\mathbb{E}[\exp\{\textrm{i}tw_{i,j}^{(l)}\frac{\phi(f_{j}^{(l-1)}(x,n))}{n^{1/\alpha}}\,|\,\{f_{j}^{(l-1)}(x,n)\}_{j=1,\ldots,n}\}]\\
&\quad=\text{e}^{-\sigma_{b}^{\alpha}|t|^{\alpha}}\prod_{j=1}^{n}\text{e}^{-\frac{\sigma_{w}^{\alpha}}{n}|t\phi(f_{j}^{(l-1)}(x,n))|^{\alpha}}\\
&\quad=\exp\left\{-(\frac{\sigma_{w}^{\alpha}}{n}\sum_{j=1}^{n}|\phi(f_{j}^{(l-1)}(x,n))|^{\alpha}+\sigma_{b}^{\alpha})|t|^{\alpha}\right\},
\end{align*}
i.e., 
\begin{equation}\label{eq:directing_measure_k_1}
f_{i}^{(l)}(x,n)\,|\,\{f_{j}^{(l-1)}(x,n)\}_{j=1,\ldots,n}\overset{\text{d}}{=}S_{\alpha,\left(\frac{\sigma_{w}^{\alpha}}{n}\sum_{j=1}^{n}|\phi(f_{j}^{(l-1)}(x,n))|^{\alpha}+\sigma_{b}^{\alpha}\right)^{1/\alpha}}.
\end{equation}
\end{itemize}
We show that, as $n\rightarrow+\infty$,
\begin{equation}\label{eq:th}
{f}_{i}^{(l)}(x,n)\overset{w}{\longrightarrow} \text{St}(\alpha,\sigma(l)),
\end{equation}
and we determine the expression of $\sigma(l)$. 

\subsection{Asymptotics for the \texorpdfstring{$i$}{i}-th coordinate}

It comes from \eqref{eq:fl} that, for every fixed $l$ and for every fixed $n$ the sequence $({f}_i^{(l)}(n,x))_{i\geq1}$ is exchangeable. In particular, let $p_{n}^{(l)}$ denote the directing (random) probability measure of the exchangeable sequence $({f}_i^{(l)}(n,x))_{i\geq1}$. That is, by de Finetti representation theorem, conditionally to $p_{n}^{(l)}$ the ${f}_i^{(l)}(n,x)$'s are iid as $p_{n}^{(l)}$. Now, consider the induction hypothesis that, as $n\rightarrow+\infty$
\begin{equation*}
p_{n}^{(l-1)}\stackrel{w}{\longrightarrow}q^{(l-1)},
\end{equation*}
with $q^{(l-1)}$ being $\text{St}(\alpha,\sigma(l-1))$, and the parameter $\sigma(l-1)$ will be specified. Therefore, we can write the following expression
\begin{align}\label{eq:espan}
\mathbb{E}[\text{e}^{\textrm{i}t{f}_i^{(l)}(x,n)}]&=\mathbb{E}\left[\exp\left\{-|t|^\alpha\left(\frac{\sigma_{w}^\alpha}{n}\sum_{j=1}^n|\phi({f}_j^{(l-1)}(x,n))|^\alpha+\sigma^{\alpha}_{b}\right)\right\}\right]\\
&=\notag\exp\left\{-|t|^{\alpha}\sigma_{b}^{\alpha}\right\}\mathbb{E}\left[\exp\left\{-|t|^\alpha\frac{\sigma_{w}^\alpha}{n}\sum_{j=1}^n|\phi({f}_j^{(l-1)}(x,n))|^\alpha\right\}\right]\\
&=\notag\exp\left\{-|t|^{\alpha}\sigma_{b}^{\alpha}\right\}\mathbb{E}\left[\left(\int \exp\left\{-|t|^\alpha\frac{\sigma_{w}^\alpha}{n}|\phi({f})|^\alpha\right\}p_{n}^{(l-1)}(\text{d}{f})\right)^n\right].
\end{align}
Hereafter we show the limiting behaviour \eqref{eq:th}. In order to prove this limiting behaviour, we will prove:
\begin{itemize}
	\item[L1)] for each $l \geq 2$ $\text{Pr}[p_{n}^{(l-1)}\in I]=1$, with $I=\{p:\int |\phi({f})|^\alpha p(\text{d}{f})<+\infty\}$;
	\item[L1.1)] for each $l \geq 2$ there exists $\epsilon >0$ such that $\sup_n \mathbb{E}[|\phi({f}_i^{(l-1)}(x,n))|^{\alpha+\epsilon}\mid p_{n}^{(l-2)}]<+\infty$;
	\item[L2)] $\int |\phi({f})|^\alpha p_{n}^{(l-1)}(\text{d}{f})\stackrel{p}{\longrightarrow} \int |\phi({f})|^\alpha q^{(l-1)}(\text{d}{f})$, as $n\rightarrow +\infty$;
	\item[L3)]  $\int |\phi({f})|^\alpha [1-e^{-|t|^\alpha\frac{\sigma^{\alpha}_{w}}{n} |\phi({f})|^\alpha}]p_{n}^{(l-1)}(\text{d}{f})\stackrel{p}{\longrightarrow} 0$, as $n\rightarrow +\infty$.
\end{itemize}

\subsubsection{Proof of L1)}

The proof of L1) follows by induction. In particular, L1) is true for the envelope condition \eqref{eq:le}, i.e., 
\begin{align*}
\mathbb{E}[|\phi({f}_i^{(1)}(x))|^\alpha]&\leq \mathbb{E}[(a+b|{f}_i^{(1)}(x)|^{\beta})^{\gamma\alpha}]\\
&\leq\mathbb{E}[a^{\gamma\alpha}+b^{\gamma\alpha}|{f}_{i}^{(1)}(x)|^{\beta\gamma\alpha}]\\
&= a^{\gamma\alpha}+b^{\gamma\alpha}\mathbb{E}[|{f}_{i}^{(1)}(x)|^{\beta\gamma\alpha}]\\
&<+\infty,
\end{align*}
since $\alpha\beta\gamma<\alpha$. Now, assuming that L1) is true for $(l-2)$, we prove that it is true for $(l-1)$. Again from \eqref{eq:le}, one has
\begin{align*}
&\mathbb{E}[|\phi({f}_i^{(l-1)}(x,n))|^\alpha\,|\, \{{f}_j^{(l-2)}(x,n)\}_{j=1,\ldots,n}]\\
&\quad\leq\mathbb{E}[(a+b|{f}_i^{(l-1)}(x,n)|^{\beta})^{\gamma\alpha}\,|\,\{{f}_j^{(l-2)}(x,n)\}_{j=1,\ldots,n}]\\
&\quad\leq \mathbb{E}[a^{\gamma\alpha}+b^{\gamma\alpha}\ |{f}_i^{(l-1)}(x,n)|^{\beta\gamma\alpha}\,|\, \{{f}_j^{(l-2)}(x,n)\}_{j=1,\ldots,n}]\\
&\quad=a^{\gamma\alpha}+b^{\gamma\alpha}\mathbb{E}[|{f}_i^{(l-1)}(x,n)|^{\beta\gamma\alpha}\,|\, \{{f}_j^{(l-2)}(x,n)\}_{j=1,\ldots,n}]\\
&\quad\leq a^{\gamma\alpha}+b^{\gamma\alpha}\mathbb{E}[|S_{\alpha,1}|^{\alpha\beta\gamma}]\left(\frac{\sigma_{w}^{\alpha}}{n}\sum_{j=1}^n |\phi({f}_j^{(l-2)}(x,n))|^\alpha+\sigma_{b}^{\alpha}\right)^{\beta\gamma}.
\end{align*}
Thus, since $\beta <\gamma^{-1}$,
\begin{align*}
&\mathbb{E}[|\phi({f}_i^{(l-1)}(x,n))|^\alpha\mid p^{(l-2)}(x,n)]\\
&\quad=\mathbb{E}[\mathbb{E}[|\phi({f}_i^{(l-1)}(x,n))|^\alpha\,|\, \{{f}_j^{(l-2)}(x,n)\}_{j=1,\ldots,n}]\,|\,p_{n}^{(l-2)}]\\
&\quad\leq a^{\gamma\alpha}+b^{\gamma\alpha}\mathbb{E}[|S_{\alpha,1}|^{\alpha\beta\gamma}]\mathbb{E}\left[\left(\frac{\sigma_{w}^{\alpha}}{n}\sum_{j=1}^n |\phi({f}_j^{(l-2)}(x,n))|^\alpha+\sigma_{b}^{\alpha}\right)^{\beta\gamma}\mid p_{n}^{(l-2)}\right]\\
&\quad\leq a^{\gamma\alpha}+b^{\gamma\alpha}\mathbb{E}[|S_{\alpha,1}|^{\alpha\beta\gamma}]\left(\mathbb{E}\left[\frac{\sigma_{w}^{\alpha}}{n}\sum_{j=1}^n |\phi({f}_j^{(l-2)}(x,n))|^\alpha+\sigma_{b}^{\alpha}\mid p_{n}^{(l-2)}\right]\right)^{\beta\gamma}\\
&\quad\leq a^{\gamma\alpha}+b^{\gamma\alpha}\mathbb{E}[|S_{\alpha,1}|^{\alpha\beta\gamma}]\left(\sigma_{b}^{\alpha}+\sigma_{w}^{\alpha}\int |\phi({f})|^\alpha p_{n}^{(l-2)}(\text{d}{f})\right)^{\beta\gamma}\\
&\quad<+\infty.
\end{align*}

\subsubsection{Proof of L1.1)}

The proof of L1.1) follows by induction, and along lines similar to the proof of L1). In particular, let $\epsilon$ be such that $\beta\gamma(\alpha+\epsilon)/\alpha<1$ and $\gamma(\alpha+\epsilon)<1$. It exists since $\beta\gamma<1$ and $\gamma\alpha<1$. For $l=2$,
\begin{align*}
\mathbb{E}(|\phi({f}_i^{(1)}(x))|^{\alpha+\epsilon})&\leq \mathbb{E}[(a+b|{f}_i^{(1)}(x)|^{\beta})^{\gamma(\alpha+\epsilon)}]\\
&\leq\mathbb{E}[a^{\gamma(\alpha+\epsilon)}+b^{\gamma(\alpha+\epsilon)}|{f}_i^{(1)}(x)|^{(\alpha+\epsilon)\gamma\beta}]\\
&=a^{\gamma(\alpha+\epsilon)}+b^{\gamma(\alpha+\epsilon)} \mathbb{E}[|{f}_i^{(1)}(x)|^{(\alpha+\epsilon)\gamma\beta}]\\
&<+\infty,
\end{align*}
since $(\alpha+\epsilon)\beta\gamma<\alpha$. This follows along lines similar to those applied in the previous subsection. Moreover the bound is uniform with respect to $n$ since the law is invariant with respect to $n$. Now, assume that L1.1) is true for $(l-2)$. Then, we can write the following inequality
\begin{align*}
&\mathbb{E}[|\phi({f}_i^{(l-1)}(x,n))|^{\alpha+\epsilon}\mid p_n^{(l-2)}]\\
&\quad\leq a^{\gamma(\alpha+\epsilon)}+ b^{\gamma(\alpha+\epsilon)}\mathbb{E}[|S_{\alpha,1}|^{\beta\gamma(\alpha+\epsilon)}]\left(\mathbb{E}\left[\frac{\sigma_{w}^{\alpha}}{n}\sum_{j=1}^n |\phi({f}_j^{(l-2)}(x,n))|^\alpha+\sigma_{b}^{\alpha}\mid p_{n}^{(l-2)}\right]\right)^{\beta\gamma(\alpha+\epsilon)/\alpha}\\
&\quad\leq a^{\gamma(\alpha+\epsilon)}+ b^{\gamma(\alpha+\epsilon)}\mathbb{E}[|S_{\alpha,1}|^{\beta\gamma(\alpha+\epsilon)}]\left(\sigma^{\alpha}_{b}+\sigma^{\alpha}_{w}\int |\phi({f})|^\alpha p_{n}^{(l-2)}(\text{d}{f})\right)^{\beta\gamma(\alpha+\epsilon)/\alpha}.
\end{align*}
Thus,
\begin{align*}
&\sup_n \mathbb{E}[|\phi({f}_i^{(l-1)}(x,n))|^{\alpha+\epsilon}\mid p_{n}^{(l-2)}]\\
&\quad\leq  a^{\gamma(\alpha+\epsilon)}+ b^{\gamma(\alpha+\epsilon)}\mathbb{E}[|S_{\alpha,1}|^{\beta\gamma(\alpha+\epsilon)}]\left(\sigma^{\alpha}_{b}+\sigma^{\alpha}_{w}\sup_n\left[\int |\phi({f})|^\alpha p_{n}^{(l-2)}(\text{d}{f})\right]\right)^{\beta\gamma(\alpha+\epsilon)/\alpha}\\
&\quad<+\infty.
\end{align*}

\subsubsection{Proof of L2)}

By the induction hypothesis, $p_n^{(l-1)}$ converges to $p^{(l-1)}$ in distribution with respect to the weak topology. Since the limit law is degenerate (on $p^{(l-1)}$), then for every subsequence $(n')$ there exists a subsequence $(n'')$ such that $p_{n''}^{(l-1)}$ converges a.s. By the induction hypothesis, $p^{(l-1)}$ is absolutely continuous with respect to the Lebesgue measure. Since $|\phi|^\alpha$ is a.s. continuous and, by L1.1), uniformly  integrable with respect to $(p_n^{(l-1)})$, then we can write the following
\[
\int |\phi({f})|^\alpha p_{n''}^{(l-1)}(\text{d}{f})\longrightarrow \int |\phi({f})|^\alpha q^{(l-1)}(\text{d}{f})\quad a.s.
\]
Thus, $n\rightarrow+\infty$
\[
\int |\phi({f})|^\alpha p_{n}^{(l-1)}(\text{d}{f})\stackrel{p}{\longrightarrow}\int |\phi({f})|^\alpha q^{(l-1)}(\text{d}{f}).
\]

\subsubsection{Proof of L3)}

Let $\epsilon$ be as in L1.1), and let $p=(\alpha+\epsilon)/\alpha$ and $q=(\alpha+\epsilon)/\epsilon$. Then $1/p+1/q=1$. Thus, by Holder inequality
\begin{align*}
&\int |\phi({f})|^\alpha [1-e^{-|t|^\alpha\frac{\sigma^{\alpha}_{w}}{n} |\phi({f})|^\alpha}]p_{n}^{(l-1)}(\text{d}{f})\\
&\quad\leq \left(\int |\phi({f})|^{\alpha p}p_{n}^{(l-1)}(\text{d}{f})\right)^{1/p}\left(\int [1-e^{-|t|^\alpha\frac{\sigma^{\alpha}_{w}}{n} |\phi({f})|^\alpha}]^{q}p_{n}^{(l-1)}(\text{d}{f}) \right)^{1/q}.
\end{align*}
Since we defined $p=(\alpha+\epsilon)/\alpha$ and $q=(\alpha+\epsilon)/\epsilon$, i.e. we set $q>1$, then we can write the following
\[
\begin{aligned}
&\left(\int |\phi({f})|^{\alpha p}p_{n}^{(l-1)}(\text{d}{f})\right)^{1/p}\left(\int [1-e^{-|t|^\alpha\frac{\sigma^{\alpha}_{w}}{n} |\phi({f})|^\alpha}]^{q}p_{n}^{(l-1)}(\text{d}{f}) \right)^{1/q}\\
&\quad \leq \sup_n \left[\left(\int |\phi({f})|^{\alpha +\epsilon}p_{n}^{(l-1)}(\text{d}{f})\right)^{1/p}\right]\left(\int [1-e^{-|t|^\alpha\frac{\sigma^{\alpha}_{w}}{n} |\phi({f})|^\alpha}]p_{n}^{(l-1)}(\text{d}{f}) \right)^{1/q}\\
&\quad \leq \sup_n \left[\left(\int |\phi({f})|^{\alpha +\epsilon}p_{n}^{(l-1)}(\text{d}{f})\right)^{1/p}\right]\left(|t|^{\alpha}\frac{\sigma^{\alpha}_{w}}{n}\int|\phi({f})|^\alpha p_{n}^{(l-1)}(\text{d}{f}) \right)^{1/q}\longrightarrow 0,
\end{aligned}
\]
as $n\rightarrow\infty$, by L1.1).

\subsubsection{Combine L1), L2) and L3)}

We combine L1), L2) and L3) to prove the large $n$  behavior of the $i$-th coordinate $n^{-1/\alpha}f_{i}(x,n)$. From \eqref{eq:espan}
\begin{align*}
&\mathbb{E}[\text{e}^{\textrm{i}t{f}_i^{(l)}(x,n)}]\\
&\quad=\notag\exp\left\{-|t|^{\alpha}\sigma_{b}^{\alpha}\right\}\mathbb{E}\left[\left(\int \exp\left\{-|t|^\alpha\frac{\sigma_{w}^\alpha}{n}|\phi({f})|^\alpha\right\}p_{n}^{(l-1)}(\text{d}{f})\right)^n\right]\\
&\quad=\notag\exp\left\{-|t|^{\alpha}\sigma_{b}^{\alpha}\right\}\mathbb{E}\left[\mathbbm{1}_{\{(p_{n}^{(l-1)}\in I)\}}\left(\int \exp\left\{-|t|^\alpha\frac{\sigma_{w}^\alpha}{n}|\phi({f})|^\alpha\right\}p_{n}^{(l-1)}(\text{d}{f})\right)^n\right].
\end{align*}
Then, by Lagrange theorem, there exists a value $\theta_{n}\in[0,1]$ such that the following equality holds true
\begin{equation*}
1-\exp\left\{-|t|^\alpha\frac{\sigma_{w}^\alpha}{n}|\phi({f})|^\alpha\right\}=|t|^\alpha\frac{\sigma_{w}^\alpha}{n}|\phi({f})|^\alpha\exp\left\{-\theta_{n}|t|^\alpha\frac{\sigma_{w}^\alpha}{n}|\phi({f})|^\alpha\right\},
\end{equation*}
thus
\begin{align*}
&\exp\left\{-|t|^\alpha\frac{\sigma_{w}^\alpha}{n}|\phi({f})|^\alpha\right\}\\
&\quad=1-|t|^\alpha\frac{\sigma_{w}^\alpha}{n}|\phi({f})|^\alpha\exp\left\{-\theta_{n}|t|^\alpha\frac{\sigma_{w}^\alpha}{n}|\phi({f})|^\alpha\right\}\\
&\quad=1-|t|^\alpha\frac{\sigma_{w}^\alpha}{n}|\phi({f})|^\alpha+|t|^\alpha\frac{\sigma_{w}^\alpha}{n}|\phi({f})|^{\alpha}\left(1-\exp\left\{-\theta_{n}|t|^\alpha\frac{\sigma_{w}^\alpha}{n}|\phi({f})|^\alpha\right\}\right).
\end{align*}
Now, since
\begin{align*}
0&\leq \int|\phi({f})|^{\alpha}[1-e^{-\theta_{n}|t|^\alpha\frac{\sigma^{\alpha}_{w}}{n} |\phi({f})|^\alpha}]p_{n}^{(l-1)}(\text{d}{f})\\
&\leq \int|\phi({f})|^{\alpha}[1-e^{-|t|^\alpha\frac{\sigma^{\alpha}_{w}}{n} |\phi({f})|^\alpha}]p_{n}^{(l-1)}(\text{d}{f}),
\end{align*}
then 
\begin{align*}
&\mathbb{E}[\text{e}^{\textrm{i}t{f}_i^{(l)}(x,n)}]\\
&\quad\leq\notag\mathbb{E}[\exp\left\{-|t|^{\alpha}\sigma_{b}^{\alpha}\right\}]\mathbb{E}\left[\mathbbm{1}_{\{(p_{n}^{(l-1)}\in I)\}}\left(1-|t|^{\alpha}\frac{\sigma^{\alpha}_{w}}{n}\int|\phi({f})|^{\alpha}p_{n}^{(l-1)}(\text{d}{f})\right.\right.\\
&\quad\quad\quad\quad\quad\quad\quad\quad\quad\quad\quad\quad\quad\quad\quad+\left.\left.|t|^{\alpha}\frac{\sigma^{\alpha}_{w}}{n}\int|\phi({f})|^{\alpha}[1-e^{-|t|^\alpha\frac{\sigma^{\alpha}_{w}}{n} |\phi({f})|^\alpha}]p_{n}^{(l-1)}(\text{d}{f})\right)^{n}\right].
\end{align*}
Thus, by using the definition of the exponential function, i.e. $\text{e}^{x}=\lim_{n\rightarrow+\infty}(1+x/n)^{n}$, and L1)-L3) we have
\begin{align*}
&\mathbb{E}[\text{e}^{\textrm{i}t{f}_i^{(l)}(x,n)}]\rightarrow e^{-|t|^\alpha[\sigma^{\alpha}_{b}+\sigma^\alpha_{w} \int |\phi({f})|^\alpha q^{(l-1)}(\text{d}{f})]},
\end{align*}
as $n\rightarrow+\infty$. That is, we proved that the large $n$ limiting distribution of ${f}_i^{(l)}(x,n)$ is $\text{St}(\alpha,\sigma(l))$, where
\begin{equation*}
\sigma(l)=\left(\sigma^{\alpha}_{b}+\sigma^\alpha_{w} \int |\phi({f})|^\alpha q^{(l-1)}(\text{d}{f})\right)^{1/\alpha}.
\end{equation*}

\newpage

\section{Large width asymptotics: \texorpdfstring{$k\geq1$}{k>=1}} \label{sec:asympt_k}

We now consider the case with of a $k$ inputs, each one being a real-valued vector of dimension $I$. We represent this generic case with a $I\times k$ input matrix $\boldsymbol{X}$. Let $\boldsymbol{1}_{r}$ denote a vector of dimension $k \times 1$ with $1$ in the $r$-the entry and $0$ elsewhere, and $\boldsymbol{1}$ denote a vector of dimension $k\times 1$ of $1$'s. If $\boldsymbol{x}_{j}$ denotes the $j$-th row of the input matrix, then we can write
\begin{equation*}
f_{i}^{(1)}(\boldsymbol{X},n)=\sum_{j=1}^{I}w_{i,j}^{(1)}\boldsymbol{x}_{j}+b_{i}^{(1)}\boldsymbol{1},
\end{equation*}
and
\begin{equation*}
f_{i}^{(l)}(\boldsymbol{X},n)=\frac{1}{n^\alpha}\sum_{j=1}^{n}w_{i,j}^{(l)}(\phi\circ f_{j}^{(l-1)}(\boldsymbol{X},n))+b_{i}^{(l)}\boldsymbol{1}
\end{equation*}
for $l=2,\ldots,D$, $i=1,\ldots,n$ where we denote with $\circ$ element-wise application. Note that $f_{i}^{(l)}(\boldsymbol{X},n)$ is a random vector of dimension $k\times1$, and we denote the $r$-th component of this vector by $f_{i,r}^{(l)}(\boldsymbol{X},n)$, namely $f_{i,r}^{(l)}(\boldsymbol{X},n)=\boldsymbol{1}_{r}^{T}f_{i}^{(l)}(\boldsymbol{X},n)$. Then, by means of \cref{eq:car_w} and \cref{eq:car_b}, we can write for $i=1 \geq 1$ and $l=2,\dots,D$:
\begin{itemize}
\item[i)]
\begin{align*}
\varphi_{f_{i}^{(1)}(\boldsymbol{X},n)}(\boldsymbol{t})&=\mathbb{E}[e^{\textrm{i}\boldsymbol{t}^{T}f_{i}^{(1)}(\boldsymbol{X},n)}]\\
&=\mathbb{E}\left[\exp\left\{\textrm{i}\boldsymbol{t}^{T}\left[\sum_{j=1}^{I}w_{i,j}^{(1)}\boldsymbol{x}_{j}+b_{i}^{(1)}\boldsymbol{1}\right]\right\}\right]\\
&=\mathbb{E}[\exp\{\textrm{i}\boldsymbol{t}^{T}b_{i}^{(1)}\boldsymbol{1}\}]\prod_{j=1}^{I}\mathbb{E}[\exp\{\textrm{i}\boldsymbol{t}^{T}w_{i,j}^{(1)}\boldsymbol{x}_{j}\}]\\
&=\text{e}^{-\sigma^{\alpha}_{b}|\boldsymbol{t}^{T}\boldsymbol{1}|^{\alpha}}\prod_{j=1}^{I}\text{e}^{-\sigma_{w}^{\alpha}|\boldsymbol{t}^{T}\boldsymbol{x}_{j}|^{\alpha}}\\
&=\exp\left\{-\sigma^{\alpha}_{b}||\boldsymbol{1}||^{\alpha}\left|\boldsymbol{t}^{T}\frac{\boldsymbol{1}}{||\boldsymbol{1}||}\right|^{\alpha}\right\}\exp\left\{-\sigma_{w}^{\alpha}\sum_{j=1}^{I}||\boldsymbol{x}_{j}||^{\alpha}\left|\boldsymbol{t}^{T}\frac{\boldsymbol{x}_{j}}{||\boldsymbol{x}_{j}||}\right|^{\alpha}\right\}\\
&=\exp\left\{-\int_{\mathbb{S}^{k-1}}|\boldsymbol{t}^{T}\boldsymbol{s}|^{\alpha}\left(||\sigma_{b}\boldsymbol{1}||^{\alpha}\delta_{\frac{\boldsymbol{1}}{||\boldsymbol{1}||}}+\sum_{j=1}^{I}||\sigma_{w}\boldsymbol{x}_{j}||^{\alpha}\delta_{\frac{\boldsymbol{x}_{j}}{||\boldsymbol{x}_{j}||}}\right)(\text{d}\boldsymbol{s})\right\},
\end{align*}
i.e.,
\begin{equation*}
f_{i}^{(1)}(\boldsymbol{X})\overset{\text{d}}{=}S_{\alpha,\Gamma^{(1)}}
\end{equation*}
with
\begin{equation*}
\Gamma^{(1)}=||\sigma_{b}\boldsymbol{1}||^{\alpha}\delta_{\frac{\boldsymbol{1}}{||\boldsymbol{1}||}}+\sum_{j=1}^{I}||\sigma_{w}\boldsymbol{x}_{j}||^{\alpha}\delta_{\frac{\boldsymbol{x}_{j}}{||\boldsymbol{x}_{j}||}};
\end{equation*}
observe that we can also determine the marginal distributions of $f_{i}^{(1)}(\boldsymbol{X})$. From \cref{eq:mar}, \cref{eq:up_s},
\begin{equation*}
f_{i,r}^{(1)}(\boldsymbol{X})\sim \text{St}(\alpha,\sigma^{(1)}(r)),
\end{equation*}
with 
\begin{equation*}
\sigma^{(1)}(r)=\left(\int_{\mathbb{S}^{k-1}}|\boldsymbol{1}_{r}^{T}\boldsymbol{s}|^{\alpha}\Gamma^{(1)}(\text{d}\boldsymbol{s})\right)^{1/\alpha}
\end{equation*}
\item[ii)]
\begin{align*}
&\varphi_{f_{i}^{(l)}(\boldsymbol{X},n)\,|\,\{f_{j}^{(l-1)}(\boldsymbol{X},n)\}_{j=1,\ldots,n}}(\boldsymbol{t})\\
&\quad=\mathbb{E}[e^{\textrm{i}\boldsymbol{t}^{T}f_{i}^{(l)}(\boldsymbol{X},n)}\,|\,\{f_{j}^{(l-1)}(\boldsymbol{X},n)\}_{j=1,\ldots,n}]\\
&\quad=\mathbb{E}\left[\exp\left\{\textrm{i}\boldsymbol{t}^{T}\left[\frac{1}{n^{1/\alpha}}\sum_{j=1}^{n}w_{i,j}^{(l)}(\phi\circ f_{j}^{(l-1)}(\boldsymbol{X},n))+b_{i}^{(l)}\boldsymbol{1}\right]\right\}\,|\,\{f_{j}^{(l-1)}(\boldsymbol{X},n)\}_{j=1,\ldots,n}\right]\\
&\quad=\mathbb{E}[\exp\{\textrm{i}\boldsymbol{t}^{T}b_{i}^{(l)}\boldsymbol{1}\}]\prod_{j=1}^{n}\mathbb{E}[\exp\{\textrm{i}\boldsymbol{t}^{T}w_{i,j}^{(l)}\frac{\phi\circ f_{j}^{(l-1)}(\boldsymbol{X},n)}{n^{1/\alpha}}\,|\,\{f_{j}^{(l-1)}(\boldsymbol{X},n)\}_{j=1,\ldots,n}\}]\\
&\quad=\text{e}^{-\sigma_{b}^{\alpha}|\boldsymbol{t}^{T}\boldsymbol{1}|^{\alpha}}\prod_{j=1}^{n}\text{e}^{-\frac{\sigma_{w}^{\alpha}}{n}|\boldsymbol{t}^{T}(\phi\circ f_{j}^{(l-1)}(\boldsymbol{X},n))|^{\alpha}}\\
&\quad=\exp\left\{-\sigma^{\alpha}_{b}||\boldsymbol{1}||^{\alpha}\left|\boldsymbol{t}^{T}\frac{\boldsymbol{1}}{||\boldsymbol{1}||}\right|^{\alpha}\right\}\\
&\quad\quad\times\exp\left\{-\frac{\sigma^{\alpha}_{w}}{n}\sum_{j=1}^{n}||\phi\circ f_{j}^{(l-1)}(\boldsymbol{X},n)||^{\alpha}\left|\boldsymbol{t}^{T}\frac{\phi\circ f_{j}^{(l-1)}(\boldsymbol{X},n)}{||\phi\circ f_{j}^{(l-1)}(\boldsymbol{X},n)||}\right|^{\alpha}\right\}\\
&\quad=\exp\left\{-\int_{\mathbb{S}^{k-1}}|\boldsymbol{t}^{T}\boldsymbol{s}|^{\alpha}\left(||\sigma_{b}\boldsymbol{1}||^{\alpha}\delta_{\frac{\boldsymbol{1}}{||\boldsymbol{1}||}}+\frac{1}{n}\sum_{j=1}^{n}||\sigma_{w}(\phi\circ f_{j}^{(l-1)}(\boldsymbol{X},n))||^{\alpha}\delta_{\frac{\phi\circ f_{j}^{(l-1)}(\boldsymbol{X},n)}{||\phi\circ f_{j}^{(l-1)}(\boldsymbol{X},n)||}}\right)(\text{d}\boldsymbol{s})\right\},
\end{align*}
i.e., 
\begin{equation*}
f_{i}^{(l)}(\boldsymbol{X},n)\,|\,\{f_{j}^{(l-1)}(\boldsymbol{X},n)\}_{j=1,\ldots,n}\overset{\text{d}}{=}S_{\alpha,\Gamma^{(l)}}
\end{equation*}
with
\begin{equation*}
\Gamma^{(l)}=||\sigma_{b}\boldsymbol{1}||^{\alpha}\delta_{\frac{\boldsymbol{1}}{||\boldsymbol{1}||}}+\frac{1}{n}\sum_{j=1}^{n}||\sigma_{w}(\phi\circ f_{j}^{(l-1)}(\boldsymbol{X},n))||^{\alpha}\delta_{\frac{\phi\circ f_{j}^{(l-1)}(\boldsymbol{X},n)}{||\phi\circ f_{j}^{(l-1)}(\boldsymbol{X},n)||}};
\end{equation*}
observe that we can also determine the marginal distributions of $f_{i}^{(l)}(\boldsymbol{X})$. From \cref{eq:mar}, \cref{eq:up_s},
\begin{equation*}
f_{i,r}^{(l)}(\boldsymbol{X},n)\,|\,\{f_{j}^{(l-1)}(\boldsymbol{X},n)\}_{j=1,\ldots,n}\sim \text{St}(\alpha,\sigma^{(l)}(r)),
\end{equation*}
with
\begin{equation}\label{eq_news}
\sigma^{(l)}(r)=\left(\int_{\mathbb{S}^{k-1}}|\boldsymbol{1}_{r}^{T}\boldsymbol{s}|^{\alpha}\Gamma^{(l)}(\text{d}\boldsymbol{s})\right)^{1/\alpha}.
\end{equation}
\end{itemize}
We show that, as $n\rightarrow+\infty$,
\begin{equation}\label{eq:th1}
{f}_{i}^{(l)}(\boldsymbol{X},n)\overset{w}{\longrightarrow}\text{St}_{k}(\alpha,\Gamma(l)),
\end{equation}
and we determine the expression of $\Gamma(l)$.

\subsection{Asymptotics for the \texorpdfstring{$i$}{i}-th coordinate}

Let $p_{n}^{(l)}$ denote the directing (random) measure of the exchangeable sequence $({f}_i^{(l)}(n,\boldsymbol{X}))_{i\geq1}$. Now, consider the induction hypothesis that,  as $n\rightarrow+\infty$,
\begin{equation*}
p_{n}^{(l-1)}\overset{w}{\longrightarrow}q^{(l-1)},
\end{equation*}
with $q^{(l-1)}$ being $\text{St}(\alpha,\sigma(l-1))$, and the finite measure $\Gamma(l-1)$ will be specified. Therefore, we can write the following expression
\begin{align}\label{eq:espan1}
&\mathbb{E}[\text{e}^{\textrm{i}\boldsymbol{t}^{T}{f}_i^{(l)}(\boldsymbol{X},n)}]\\
&\notag\quad=\mathbb{E}\left[\exp\left\{-\int_{\mathbb{S}^{k-1}}|\boldsymbol{t}^{T}\boldsymbol{s}|^{\alpha}\tilde{\Gamma}^{(l)}(\text{d}\boldsymbol{s})\right\}\right]\\
&\notag\quad=\mathbb{E}\left[\exp\left\{-\int_{\mathbb{S}^{k-1}}|\boldsymbol{t}^{T}\boldsymbol{s}|^{\alpha}\left(||\sigma_{b}\boldsymbol{1}||^{\alpha}\delta_{\frac{\boldsymbol{1}}{||\boldsymbol{1}||}}\right)(\text{d}\boldsymbol{s})\right\}\right]\\
&\notag\quad\quad\times\mathbb{E}\left[\exp\left\{-\int_{\mathbb{S}^{k-1}}|\boldsymbol{t}^{T}\boldsymbol{s}|^{\alpha}\left(\frac{1}{n}\sum_{j=1}^{n}||\sigma_{w}(\phi\circ f_{j}^{(l-1)}(\boldsymbol{X},n))||^{\alpha}\delta_{\frac{\phi\circ f_{j}^{(l-1)}(\boldsymbol{X},n)}{||\phi\circ f_{j}^{(l-1)}(\boldsymbol{X},n)||}}\right)(\text{d}\boldsymbol{s})\right\}\right]\\
&\notag\quad=\mathbb{E}\left[\exp\left\{-\int_{\mathbb{S}^{k-1}}|\boldsymbol{t}^{T}\boldsymbol{s}|^{\alpha}\left(||\sigma_{b}\boldsymbol{1}||^{\alpha}\delta_{\frac{\boldsymbol{1}}{||\boldsymbol{1}||}}\right)(\text{d}\boldsymbol{s})\right\}\right]\\
&\notag\quad\quad\times\mathbb{E}\left[\left(\int\exp\left\{-\int_{\mathbb{S}^{k-1}}|\boldsymbol{t}^{T}\boldsymbol{s}|^{\alpha}\left(\frac{1}{n}||\sigma_{w}(\phi\circ{f})||^{\alpha}\delta_{\frac{\phi\circ{f}}{||\phi\circ{f}||}}\right)(\text{d}\boldsymbol{s})\right\}p_{n}^{(l-1)}(\text{d}{f})\right)^{n}\right].
\end{align}
Hereafter we show the limiting behaviour \cref{eq:th1}. In order to prove this limiting behaviour, we will prove:
\begin{itemize}
	\item[L1)] for each $l \geq 2$ $\text{Pr}[p_{n}^{(l-1)}\in I]=1$, with $I=\{p:\int ||\phi\circ{f}||^\alpha p(\text{d}{f})<+\infty\}$;
	\item[L1.1)] for each $l \geq 2$ there exists $\epsilon >0$ such that $\sup_n \mathbb{E}[||\phi\circ{f}_i^{(l-1)}(\boldsymbol{X},n)||^{\alpha+\epsilon}\mid p_{n}^{(l-2)}]<+\infty$;
	\item[L2)] $\int ||\phi\circ{f}||^\alpha p_{n}^{(l-1)}(\text{d}{f})\overset{p}{\longrightarrow} \int ||\phi\circ{f}||^\alpha q^{(l-1)}(\text{d}{f})$, as $n\rightarrow +\infty$;
	\item[L3)]  $\int ||\phi\circ{f}||^\alpha \left[1-\exp\left\{-\int_{\mathbb{S}^{k-1}}|\boldsymbol{t}^{T}\boldsymbol{s}|^{\alpha}\left(\frac{1}{n}||\sigma_{w}(\phi\circ{f})||^{\alpha}\delta_{\frac{\phi\circ{f}}{||\phi\circ{f}||}}\right)(\text{d}\boldsymbol{s})\right\}\right]p_{n}^{(l-1)}(\text{d}{f})\overset{p}{\longrightarrow} 0$, as $n\rightarrow +\infty$.
\end{itemize}

\subsubsection{Proof of L1)}

The proof of L1) follows by induction. In particular, L1) is true for the envelope condition \cref{eq:le}, i.e., 
\begin{align*}
\mathbb{E}[||\phi\circ{f}_i^{(1)}(\boldsymbol{X})||^\alpha]&\leq\mathbb{E}\left[\sum_{r=1}^{k}|\phi({f}_{i,r}^{(1)}(\boldsymbol{X}))|^{\alpha}\right]\\
&\leq \sum_{r=1}^{k}\mathbb{E}[(a+b|{f}_{i,r}^{(1)}(\boldsymbol{X})|^{\beta})^{\gamma\alpha}]\\
&\leq \sum_{r=1}^{k}\mathbb{E}[(a^{\gamma\alpha}+b^{\gamma\alpha}|{f}_{i,r}^{(1)}(\boldsymbol{X})|^{\beta\gamma\alpha})]\\
&\leq ka^{\gamma\alpha}+b^{\gamma\alpha}\sum_{r=1}^{k}\mathbb{E}[|{f}_{i,r}^{(1)}(\boldsymbol{X})|^{\beta\gamma\alpha}]\\
&<+\infty
\end{align*}
since $\alpha\beta\gamma<\alpha$. Now, assuming that L1) is true for $(l-2)$, we prove that it is true for $(l-1)$. Again from \cref{eq:le},
\begin{align*}
&\mathbb{E}[||\phi\circ{f}_i^{(l-1)}(\boldsymbol{X},n)||^\alpha\,|\, \{{f}_j^{(l-2)}(\boldsymbol{X},n)\}_{j=1,\ldots,n}]\\
&\quad\leq\mathbb{E}\left[\sum_{r=1}^{k}|\phi({f}_{i,r}^{(l-1)}(\boldsymbol{X},n))|^{\alpha}\,|\,\{{f}_j^{(l-2)}(\boldsymbol{X},n)\}_{j=1,\ldots,n}\right]\\
&\quad\leq \sum_{r=1}^{k}\mathbb{E}[(a+b\ |{f}_{i,r}^{(l-1)}(\boldsymbol{X},n)|^{\beta})^{\gamma\alpha}\,|\, \{{f}_j^{(l-2)}(\boldsymbol{X},n)\}_{j=1,\ldots,n}]\\
&\quad\leq \sum_{r=1}^{k}\mathbb{E}[(a^{\gamma\alpha}+b^{\gamma\alpha}\ |{f}_{i,r}^{(l-1)}(\boldsymbol{X},n)|^{\beta\gamma\alpha})\,|\, \{{f}_j^{(l-2)}(\boldsymbol{X},n)\}_{j=1,\ldots,n}]\\
&\quad= ka^{\gamma\alpha}+b^{\gamma\alpha}\sum_{r=1}^{k}\mathbb{E}[|{f}_{i,r}^{(l-1)}(\boldsymbol{X},n)|^{\beta\gamma\alpha}\,|\,\{{f}_j^{(l-2)}(\boldsymbol{X},n)\}_{j=1,\ldots,n}]\\
&\quad\leq ka^{\gamma\alpha}+b^{\gamma\alpha}\mathbb{E}[|S_{\alpha,1}|^{\alpha\beta\gamma}]\\
&\quad\quad\times\sum_{r=1}^{k}\left(\int_{\mathbb{S}^{k-1}}|\boldsymbol{1}_{r}^{T}\boldsymbol{s}|^{\alpha}\left(||\sigma_{b}\boldsymbol{1}||^{\alpha}\delta_{\frac{\boldsymbol{1}}{||\boldsymbol{1}||}}\right)(\text{d}\boldsymbol{s})\right.\\
&\left.\quad\quad\quad\quad\quad\quad\quad\quad\quad+\int_{\mathbb{S}^{k-1}}|\boldsymbol{1}_{r}^{T}\boldsymbol{s}|^{\alpha}\left(\frac{1}{n}\sum_{j=1}^{n}||\sigma_{w}(\phi\circ f_{j}^{(l-2)}(\boldsymbol{X},n))||^{\alpha}\delta_{\frac{\phi\circ f_{j}^{(l-2)}(\boldsymbol{X},n)}{||\phi\circ f_{j}^{(l-2)}(\boldsymbol{X},n)||}}\right)(\text{d}\boldsymbol{s})\right)^{\beta\gamma}.
\end{align*}

Since $\beta<\gamma^{-1}$,
\begin{align*}
&\mathbb{E}[||\phi\circ{f}_i^{(l-1)}(\boldsymbol{X},n)||^\alpha\,|\, p_{n}^{(l-2)}]\\
&\quad=\mathbb{E}[\mathbb{E}[||\phi\circ{f}_i^{(l-1)}(\boldsymbol{X},n)||^\alpha\,|\, \{{f}_j^{(l-2)}(\boldsymbol{X},n)\}_{j=1,\ldots,n}]\,|\,p_{n}^{(l-2)}]\\
&\quad\leq ka^{\gamma\alpha}+b^{\gamma\alpha}\mathbb{E}[|S_{\alpha,1}|^{\alpha\beta\gamma}]\\
&\quad\quad\times\sum_{r=1}^{k}\mathbb{E}\left[\left(\int_{\mathbb{S}^{k-1}}|\boldsymbol{1}_{r}^{T}\boldsymbol{s}|^{\alpha}\left(||\sigma_{b}\boldsymbol{1}||^{\alpha}\delta_{\frac{\boldsymbol{1}}{||\boldsymbol{1}||}}\right)(\text{d}\boldsymbol{s})\right.\right.\\
&\quad\quad\quad\quad\quad\left.\left.+\int_{\mathbb{S}^{k-1}}|\boldsymbol{1}_{r}^{T}\boldsymbol{s}|^{\alpha}\left(\frac{1}{n}\sum_{j=1}^{n}||\sigma_{w}(\phi\circ f_{j}^{(l-2)}(\boldsymbol{X},n))||^{\alpha}\delta_{\frac{\phi\circ f_{j}^{(l-2)}(\boldsymbol{X},n)}{||\phi\circ f_{j}^{(l-2)}(\boldsymbol{X},n)||}}\right)(\text{d}\boldsymbol{s})\right)^{\beta\gamma}\,|\,p_{n}^{(l-2)}\right]\\
&\quad\leq ka^{\gamma\alpha}+b^{\gamma\alpha}\mathbb{E}[|S_{\alpha,1}|^{\alpha\beta\gamma}]\\
&\quad\quad\times\sum_{r=1}^{k}\left(\mathbb{E}\left[\int_{\mathbb{S}^{k-1}}|\boldsymbol{1}_{r}^{T}\boldsymbol{s}|^{\alpha}\left(||\sigma_{b}\boldsymbol{1}||^{\alpha}\delta_{\frac{\boldsymbol{1}}{||\boldsymbol{1}||}}\right)(\text{d}\boldsymbol{s})\right.\right.\\
&\quad\quad\quad\quad\quad\left.\left.+\int_{\mathbb{S}^{k-1}}|\boldsymbol{1}_{r}^{T}\boldsymbol{s}|^{\alpha}\left(\frac{1}{n}\sum_{j=1}^{n}||\sigma_{w}(\phi\circ f_{j}^{(l-2)}(\boldsymbol{X},n))||^{\alpha}\delta_{\frac{\phi\circ f_{j}^{(l-2)}(\boldsymbol{X},n)}{||\phi\circ f_{j}^{(l-2)}(\boldsymbol{X},n)||}}\right)(\text{d}\boldsymbol{s})\,|\,p_{n}^{(l-2)}\right]\right)^{\beta\gamma}\\
&\quad\leq ka^{\gamma\alpha}+b^{\gamma\alpha}\mathbb{E}[|S_{\alpha,1}|^{\alpha\beta\gamma}]\\
&\quad\quad\times\sum_{r=1}^{k}\left(\int_{\mathbb{S}^{k-1}}|\boldsymbol{1}_{r}^{T}\boldsymbol{s}|^{\alpha}\left(||\sigma_{b}\boldsymbol{1}||^{\alpha}\delta_{\frac{\boldsymbol{1}}{||\boldsymbol{1}||}}\right)(\text{d}\boldsymbol{s})\right.\\
&\quad\quad\quad\quad\quad\quad\quad\quad\quad\quad\left.+\int_{\mathbb{S}^{k-1}}|\boldsymbol{1}_{r}^{T}\boldsymbol{s}|^{\alpha}\left(\int||\sigma_{w}(\phi\circ{f})||^{\alpha}\delta_{\frac{\phi\circ{f}}{||\phi\circ{f}||}}p_{n}^{(l-2)}(\text{d}{f})\right)(\text{d}\boldsymbol{s})\right)^{\beta\gamma}\\
&\quad<+\infty.
\end{align*}

\subsubsection{Proof of L1.1)}

The proof of L1.1) follows by induction, and along lines similar to the proof of L1). In particular, let $\epsilon$ be such that $\beta\gamma(\alpha+\epsilon)/\alpha<1$ and $\gamma(\alpha+\epsilon)<1$. It exists since $\beta\gamma<1$ and $\gamma\alpha<1$. For $l=2$, we can find $C(k) > 0$ finite such that:
\begin{align*}
\mathbb{E}(||\phi\circ{f}_i^{(1)}(\boldsymbol{X})||^{\alpha+\epsilon})&\leq \mathbb{E}\left[C(k) \sum_{r=1}^{k}|\phi({f}_{i,r}^{(1)}(\boldsymbol{X}))|^{\alpha+\epsilon}\right]\\
&\leq C(k) \sum_{r=1}^{k}\mathbb{E}[(a+b|{f}_{i,r}^{(1)}(\boldsymbol{X})|^{\beta})^{\gamma(\alpha+\epsilon)}]\\
&\leq C(k) \sum_{r=1}^{k}\mathbb{E}[(a^{\gamma(\alpha+\epsilon)}+b^{\gamma(\alpha+\epsilon)}|{f}_{i,r}^{(1)}(\boldsymbol{X})|^{\beta\gamma(\alpha+\epsilon)})]\\
&= C(k) \left(ka^{\gamma(\alpha+\epsilon)}+b^{\gamma(\alpha+\epsilon)}\sum_{r=1}^{k}\mathbb{E}[|{f}_{i,r}^{(1)}(\boldsymbol{X})|^{(\alpha+\epsilon)\gamma\beta}]\right)\\
&<+\infty,
\end{align*}
since $(\alpha+\epsilon)\beta\gamma<\alpha$. This follows along lines similar to those applied in the previous subsection. Moreover the bound is uniform with respect to $n$ since the law is invariant with respect to $n$. Let us assume that L1.1) is true for $(l-2)$. Then we can write the following
\begin{align*}
&\mathbb{E}[||\phi\circ{f}_i^{(l-1)}(\boldsymbol{X},n)||^{\alpha+\epsilon}\mid \{f_{j}^{(l-2)}(x,n)\}_{j=1,\ldots,n}]\\
&\quad\leq C(k) \Bigg\{ka^{\gamma(\alpha+\epsilon)}+b^{\gamma(\alpha+\epsilon)}\mathbb{E}[|S_{\alpha,1}|^{\beta\gamma(\alpha+\epsilon)}]\\
&\quad\quad\times\sum_{r=1}^{k}\left(\mathbb{E}\left[\int_{\mathbb{S}^{k-1}}|\boldsymbol{1}_{r}^{T}\boldsymbol{s}|^{\alpha}\left(||\sigma_{b}\boldsymbol{1}||^{\alpha}\delta_{\frac{\boldsymbol{1}}{||\boldsymbol{1}||}}\right)(\text{d}\boldsymbol{s})\right.\right.\\
&\quad\quad\quad\quad\quad\left.\left.+\int_{\mathbb{S}^{k-1}}|\boldsymbol{1}_{r}^{T}\boldsymbol{s}|^{\alpha}\left(\frac{1}{n}\sum_{j=1}^{n}||\sigma_{w}(\phi\circ f_{j}^{(l-2)}(\boldsymbol{X},n))||^{\alpha}\delta_{\frac{\phi\circ f_{j}^{(l-2)}(\boldsymbol{X},n)}{||\phi\circ f_{j}^{(l-2)}(\boldsymbol{X},n)||}}\right)(\text{d}\boldsymbol{s})\,|\,p_{n}^{(l-2)}\right]\right)^{\beta\gamma(\alpha+\epsilon)/\alpha} \Bigg\}\\
&\quad\leq C(k) \Bigg\{ka^{\gamma(\alpha+\epsilon)}+b^{\gamma(\alpha+\epsilon)}\mathbb{E}[|S_{\alpha,1}|^{\beta\gamma(\alpha+\epsilon)}]\\
&\quad\quad\times\sum_{r=1}^{k}\left(\int_{\mathbb{S}^{k-1}}|\boldsymbol{1}_{r}^{T}\boldsymbol{s}|^{\alpha}\left(||\sigma_{b}\boldsymbol{1}||^{\alpha}\delta_{\frac{\boldsymbol{1}}{||\boldsymbol{1}||}}\right)(\text{d}\boldsymbol{s})\right.\\
&\quad\quad\quad\quad\quad\quad\quad\quad\quad\quad\left.+\int_{\mathbb{S}^{k-1}}|\boldsymbol{1}_{r}^{T}\boldsymbol{s}|^{\alpha}\left(\int||\sigma_{w}(\phi\circ{f})||^{\alpha}\delta_{\frac{\phi\circ{f}}{||\phi\circ{f}||}}p_{n}^{(l-2)}(\text{d}{f})\right)(\text{d}\boldsymbol{s})\right)^{\beta\gamma(\alpha+\epsilon)/\alpha}\Bigg\}
\end{align*}
Thus,
\begin{align*}
&\sup_n \mathbb{E}[||\phi\circ{f}_i^{(l-1)}(\boldsymbol{X},n)||^{\alpha+\epsilon}\mid p_{n}^{(l-2)}]\\
&\quad\leq C(k) \Bigg\{ka^{\gamma(\alpha+\epsilon)}+b^{\gamma(\alpha+\epsilon)}\mathbb{E}[|S_{\alpha,1}|^{\beta\gamma(\alpha+\epsilon)}]\\
&\quad\quad\times\ \sup_n \sum_{r=1}^{k}\left(\int_{\mathbb{S}^{k-1}}|\boldsymbol{1}_{r}^{T}\boldsymbol{s}|^{\alpha}\left(||\sigma_{b}\boldsymbol{1}||^{\alpha}\delta_{\frac{\boldsymbol{1}}{||\boldsymbol{1}||}}\right)(\text{d}\boldsymbol{s})\right.\\
&\quad\quad\quad\quad\quad\quad\left.+\int_{\mathbb{S}^{k-1}}|\boldsymbol{1}_{r}^{T}\boldsymbol{s}|^{\alpha}\left(\int||\sigma_{w}(\phi\circ{f})||^{\alpha}\delta_{\frac{\phi\circ{f}}{||\phi\circ{f}||}}p_{n}^{(l-2)}(\text{d}{f})\right)(\text{d}\boldsymbol{s})\right)^{\beta\gamma(\alpha+\epsilon)/\alpha} \Bigg\}\\
&\quad<+\infty.
\end{align*}

\subsubsection{Proof of L2)}

By the induction hypothesis, $p_n^{(l-1)}$ converges to $p^{(l-1)}$ in distribution with respect to the weak topology. Since the limit law is degenerate (on $p^{(l-1)}$), then for every subsequence $(n')$ there exists a subsequence $(n'')$ such that $p_{n''}^{(l-1)}$ converges a.s. By the induction hypothesis, $p^{(l-1)}$ is absolutely continuous with respect to the Lebesgue measure. Since $|\phi|^\alpha$ is a.s. continuous and, by L1.1), uniformly  integrable with respect to $(p_n^{(l-1)})$, then we can write the following
\[
\int ||\phi\circ{f}||^\alpha p_{n''}^{(l-1)}(\text{d}{f})\longrightarrow \int ||\phi\circ{f}||^\alpha q^{(l-1)}(\text{d}{f})\quad a.s.
\]
Thus, $n\rightarrow+\infty$
\[
\int ||\phi\circ{f}||^\alpha p_{n}^{(l-1)}(\text{d}{f})\overset{p}{\longrightarrow}\int ||\phi\circ{f}||^\alpha q^{(l-1)}(\text{d}{f}).
\]

\subsubsection{Proof of L3)}

Let $\epsilon$ be as in L1.1), and let $p=(\alpha+\epsilon)/\alpha$ and $q=(\alpha+\epsilon)/\epsilon$. Then $1/p+1/q=1$. Thus, by Holder inequality
\begin{align*}
&\int ||\phi\circ{f}||^\alpha\left[1-\exp\left\{-\int_{\mathbb{S}^{k-1}}|\boldsymbol{t}^{T}\boldsymbol{s}|^{\alpha}\left(\frac{1}{n}||\sigma_{w}(\phi\circ{f})||^{\alpha}\delta_{\frac{\phi\circ{f}}{||\phi\circ{f}||}}\right)(\text{d}\boldsymbol{s})\right\}\right]p_{n}^{(l-1)}(\text{d}{f})\\
&\quad\leq \left(\int ||\phi\circ{f}||^{\alpha p}p_{n}^{(l-1)}(\text{d}{f})\right)^{1/p}\\
&\quad\quad\times\left(\int \left[1-\exp\left\{-\int_{\mathbb{S}^{k-1}}|\boldsymbol{t}^{T}\boldsymbol{s}|^{\alpha}\left(\frac{1}{n}||\sigma_{w}(\phi\circ{f})||^{\alpha}\delta_{\frac{\phi\circ{f}}{||\phi\circ{f}||}}\right)(\text{d}\boldsymbol{s})\right\}\right]^{q}p_{n}^{(l-1)}(\text{d}{f}) \right)^{1/q}
\end{align*}
Since we defined $p=(\alpha+\epsilon)/\alpha$ and $q=(\alpha+\epsilon)/\epsilon$, i.e. we set $q>1$, then we can write the following
\[
\begin{aligned}
&\left(\int ||\phi\circ{f}||^{\alpha p}p_{n}^{(l-1)}(\text{d}{f})\right)^{1/p}\\
&\quad\quad\times\left(\int \left[1-\exp\left\{-\int_{\mathbb{S}^{k-1}}|\boldsymbol{t}^{T}\boldsymbol{s}|^{\alpha}\left(\frac{1}{n}||\sigma_{w}(\phi\circ{f})||^{\alpha}\delta_{\frac{\phi\circ{f}}{||\phi\circ{f}||}}\right)(\text{d}\boldsymbol{s})\right\}\right]^{q}p_{n}^{(l-1)}(\text{d}{f}) \right)^{1/q}\\
&\quad \leq \sup_n \left[\left(\int ||\phi\circ{f}||^{\alpha +\epsilon}p_{n}^{(l-1)}(\text{d}{f})\right)^{1/p}\right]\\
&\quad\quad\times\left(\int \left[1-\exp\left\{-\int_{\mathbb{S}^{k-1}}|\boldsymbol{t}^{T}\boldsymbol{s}|^{\alpha}\left(\frac{1}{n}||\sigma_{w}(\phi\circ{f})||^{\alpha}\delta_{\frac{\phi\circ{f}}{||\phi\circ{f}||}}\right)(\text{d}\boldsymbol{s})\right\}\right]p_{n}^{(l-1)}(\text{d}{f}) \right)^{1/q}\\
&\quad \leq \sup_n \left[\left(\int ||\phi\circ{f}||^{\alpha +\epsilon}p_{n}^{(l-1)}(\text{d}{f})\right)^{1/p}\right]\\
&\quad\quad\times\left(\left[\int_{\mathbb{S}^{k-1}}|\boldsymbol{t}^{T}\boldsymbol{s}|^{\alpha}\left(\frac{1}{n}||\sigma_{w}(\phi\circ{f})||^{\alpha}\delta_{\frac{\phi\circ{f}}{||\phi\circ{f}||}}\right)(\text{d}\boldsymbol{s})\right] p_{n}^{(l-1)}(\text{d}{f}) \right)^{1/q}\rightarrow 0,
\end{aligned}
\]
as $n\rightarrow\infty$, by L1.1).

\subsubsection{Combine L1), L2) and L3)}

We combine L1), L2) and L3) to prove the large $n$  behavior of the $i$-th coordinate $n^{-1/\alpha}f_{i}(x,n)$. From \cref{eq:espan}
\begin{align*}
&\mathbb{E}[\text{e}^{\textrm{i}\boldsymbol{t}^{T}{f}_i^{(l)}(\boldsymbol{X},n)}]\\
&\notag\quad=\mathbb{E}\left[\exp\left\{-\int_{\mathbb{S}^{k-1}}|\boldsymbol{t}^{T}\boldsymbol{s}|^{\alpha}\left(||\sigma_{b}\boldsymbol{1}||^{\alpha}\delta_{\frac{\boldsymbol{1}}{||\boldsymbol{1}||}}\right)(\text{d}\boldsymbol{s})\right\}\right]\\
&\notag\quad\quad\times\mathbb{E}\left[\left(\int\exp\left\{-\int_{\mathbb{S}^{k-1}}|\boldsymbol{t}^{T}\boldsymbol{s}|^{\alpha}\left(\frac{1}{n}||\sigma_{w}(\phi\circ{f})||^{\alpha}\delta_{\frac{\phi\circ{f}}{||\phi\circ{f}||}}\right)(\text{d}\boldsymbol{s})\right\}p_{n}^{(l-1)}(\text{d}{f})\right)^{n}\right].\\
&\notag\quad=\mathbb{E}\left[\exp\left\{-\int_{\mathbb{S}^{k-1}}|\boldsymbol{t}^{T}\boldsymbol{s}|^{\alpha}\left(||\sigma_{b}\boldsymbol{1}||^{\alpha}\delta_{\frac{\boldsymbol{1}}{||\boldsymbol{1}||}}\right)(\text{d}\boldsymbol{s})\right\}\right]\\
&\quad\quad\times\mathbb{E}\left[\mathbbm{1}_{\{(p_{n}^{(l-1)}\in I)\}}\left(\int\exp\left\{-\int_{\mathbb{S}^{k-1}}|\boldsymbol{t}^{T}\boldsymbol{s}|^{\alpha}\left(\frac{1}{n}||\sigma_{w}(\phi\circ{f})||^{\alpha}\delta_{\frac{\phi\circ{f}}{||\phi\circ{f}||}}\right)(\text{d}\boldsymbol{s})\right\}p_{n}^{(l-1)}(\text{d}{f})\right)^{n}\right].
\end{align*}
Then, by Lagrange theorem, there exists a value $\theta_{n}\in[0,1]$ such that the following equality holds true
\begin{align*}
&1-\exp\left\{-\int_{\mathbb{S}^{k-1}}|\boldsymbol{t}^{T}\boldsymbol{s}|^{\alpha}\left(\frac{1}{n}||\sigma_{w}(\phi\circ{f})||^{\alpha}\delta_{\frac{\phi\circ{f}}{||\phi\circ{f}||}}\right)(\text{d}\boldsymbol{s})\right\}\\
&\quad=\left(\int_{\mathbb{S}^{k-1}}|\boldsymbol{t}^{T}\boldsymbol{s}|^{\alpha}\left(\frac{1}{n}||\sigma_{w}(\phi\circ{f})||^{\alpha}\delta_{\frac{\phi\circ{f}}{||\phi\circ{f}||}}\right)(\text{d}\boldsymbol{s})\right)\\
&\quad\quad\times\exp\left\{-\theta_{n}\int_{\mathbb{S}^{k-1}}|\boldsymbol{t}^{T}\boldsymbol{s}|^{\alpha}\left(\frac{1}{n}||\sigma_{w}(\phi\circ{f})||^{\alpha}\delta_{\frac{\phi\circ{f}}{||\phi\circ{f}||}}\right)(\text{d}\boldsymbol{s})\right\},
\end{align*}
thus
\begin{align*}
&\exp\left\{-\int_{\mathbb{S}^{k-1}}|\boldsymbol{t}^{T}\boldsymbol{s}|^{\alpha}\left(\frac{1}{n}||\sigma_{w}(\phi\circ{f})||^{\alpha}\delta_{\frac{\phi\circ{f}}{||\phi\circ{f}||}}\right)(\text{d}\boldsymbol{s})\right\}\\
&\quad=1-\left(\int_{\mathbb{S}^{k-1}}|\boldsymbol{t}^{T}\boldsymbol{s}|^{\alpha}\left(\frac{1}{n}||\sigma_{w}(\phi\circ{f})||^{\alpha}\delta_{\frac{\phi\circ{f}}{||\phi\circ{f}||}}\right)(\text{d}\boldsymbol{s})\right)\\
&\quad\quad\quad\quad\times\exp\left\{-\theta_{n}\int_{\mathbb{S}^{k-1}}|\boldsymbol{t}^{T}\boldsymbol{s}|^{\alpha}\left(\frac{1}{n}||\sigma_{w}(\phi\circ{f})||^{\alpha}\delta_{\frac{\phi\circ{f}}{||\phi\circ{f}||}}\right)(\text{d}\boldsymbol{s})\right\}\\
&\quad=1-\left(\int_{\mathbb{S}^{k-1}}|\boldsymbol{t}^{T}\boldsymbol{s}|^{\alpha}\left(\frac{1}{n}||\sigma_{w}(\phi\circ{f})||^{\alpha}\delta_{\frac{\phi\circ{f}}{||\phi\circ{f}||}}\right)(\text{d}\boldsymbol{s})\right)\\
&\quad\quad\quad\quad+\left(\int_{\mathbb{S}^{k-1}}|\boldsymbol{t}^{T}\boldsymbol{s}|^{\alpha}\left(\frac{1}{n}||\sigma_{w}(\phi\circ{f})||^{\alpha}\delta_{\frac{\phi\circ{f}}{||\phi\circ{f}||}}\right)(\text{d}\boldsymbol{s})\right)\\
&\quad\quad\quad\quad\quad\times\left(1-\exp\left\{-\theta_{n}\int_{\mathbb{S}^{k-1}}|\boldsymbol{t}^{T}\boldsymbol{s}|^{\alpha}\left(\frac{1}{n}||\sigma_{w}(\phi\circ{f})||^{\alpha}\delta_{\frac{\phi\circ{f}}{||\phi\circ{f}||}}\right)(\text{d}\boldsymbol{s})\right\}\right).
\end{align*}
Now, since
\begin{align*}
0&\leq\int\int_{\mathbb{S}^{k-1}}|\boldsymbol{t}^{T}\boldsymbol{s}|^{\alpha}\left(\frac{1}{n}||\sigma_{w}(\phi\circ{f})||^{\alpha}\delta_{\frac{\phi\circ{f}}{||\phi\circ{f}||}}\right)(\text{d}\boldsymbol{s})\\
&\quad\quad\quad\quad\quad\quad\left.\times\left[1-\exp\left\{-\theta_{n}\int_{\mathbb{S}^{k-1}}|\boldsymbol{t}^{T}\boldsymbol{s}|^{\alpha}\left(\frac{1}{n}||\sigma_{w}(\phi\circ{f})||^{\alpha}\delta_{\frac{\phi\circ{f}}{||\phi\circ{f}||}}\right)(\text{d}\boldsymbol{s})\right\}\right]p_{n}^{(l-1)}(\text{d}{f})\right.\\
&\leq\int\int_{\mathbb{S}^{k-1}}|\boldsymbol{t}^{T}\boldsymbol{s}|^{\alpha}\left(\frac{1}{n}||\sigma_{w}(\phi\circ{f})||^{\alpha}\delta_{\frac{\phi\circ{f}}{||\phi\circ{f}||}}\right)(\text{d}\boldsymbol{s})\\
&\quad\quad\quad\quad\quad\quad\left.\times\left[1-\exp\left\{-\int_{\mathbb{S}^{k-1}}|\boldsymbol{t}^{T}\boldsymbol{s}|^{\alpha}\left(\frac{1}{n}||\sigma_{w}(\phi\circ{f})||^{\alpha}\delta_{\frac{\phi\circ{f}}{||\phi\circ{f}||}}\right)(\text{d}\boldsymbol{s})\right\}\right]p_{n}^{(l-1)}(\text{d}{f})\right.
\end{align*}
then
\begin{align*}
&\mathbb{E}[\text{e}^{\textrm{i}\boldsymbol{t}^{T}{f}_i^{(l)}(\boldsymbol{X},n)}]\\
&\quad\leq\mathbb{E}\left[\exp\left\{-\int_{\mathbb{S}^{k-1}}|\boldsymbol{t}^{T}\boldsymbol{s}|^{\alpha}\left(||\sigma_{b}\boldsymbol{1}||^{\alpha}\delta_{\frac{\boldsymbol{1}}{||\boldsymbol{1}||}}\right)(\text{d}\boldsymbol{s})\right\}\right]\\
&\quad\quad\times\mathbb{E}\left[\mathbbm{1}_{\{(p_{n}^{(l-1)}\in I)\}}\left(1-\int\int_{\mathbb{S}^{k-1}}|\boldsymbol{t}^{T}\boldsymbol{s}|^{\alpha}\left(\frac{1}{n}||\sigma_{w}(\phi\circ{f})||^{\alpha}\delta_{\frac{\phi\circ{f}}{||\phi\circ{f}||}}\right)(\text{d}\boldsymbol{s})p_{n}^{(l-1)}(\text{d}{f})\right.\right.\\
&\quad\quad\quad+\int\int_{\mathbb{S}^{k-1}}|\boldsymbol{t}^{T}\boldsymbol{s}|^{\alpha}\left(\frac{1}{n}||\sigma_{w}(\phi\circ{f})||^{\alpha}\delta_{\frac{\phi\circ{f}}{||\phi\circ{f}||}}\right)(\text{d}\boldsymbol{s})\\
&\quad\quad\quad\quad\quad\quad\left.\left.\left.\times\left[1-\exp\left\{-\int_{\mathbb{S}^{k-1}}|\boldsymbol{t}^{T}\boldsymbol{s}|^{\alpha}\left(\frac{1}{n}||\sigma_{w}(\phi\circ{f})||^{\alpha}\delta_{\frac{\phi\circ{f}}{||\phi\circ{f}||}}\right)(\text{d}\boldsymbol{s})\right\}\right]p_{n}^{(l-1)}(\text{d}{f})\right.\right)^{n}\right].
\end{align*}
Thus, by using the definition of the exponential function, i.e. $\text{e}^{x}=\lim_{n\rightarrow+\infty}(1+x/n)^{n}$, and L1)-L3) we have
\begin{align*}
&\mathbb{E}[\text{e}^{\textrm{i}\boldsymbol{t}^{T}{f}_i^{(l)}(\boldsymbol{X},n)}]\\
&\quad\rightarrow\exp\left\{-\int_{\mathbb{S}^{k-1}}|\boldsymbol{t}^{T}\boldsymbol{s}|^{\alpha}\left(||\sigma_{b}\boldsymbol{1}||^{\alpha}\delta_{\frac{\boldsymbol{1}}{||\boldsymbol{1}||}}\right)(\text{d}\boldsymbol{s})\right\}\\
&\quad\quad\times\exp\left\{-\int\int_{\mathbb{S}^{k-1}}|\boldsymbol{t}^{T}\boldsymbol{s}|^{\alpha}\left(||\sigma_{w}(\phi\circ{f})||^{\alpha}\delta_{\frac{\phi\circ{f}}{||\phi\circ{f}||}}\right)(\text{d}\boldsymbol{s})q^{(l-1)}(\text{d}{f})\right\}
\end{align*}
as $n\rightarrow+\infty$. That is, we proved that the large $n$ limiting distribution of ${f}_i^{(l)}(x,n)$ is $\text{St}_{k}(\alpha,\Gamma(l))$, where
\begin{equation}\label{eq:parametro}
\Gamma(l)=||\sigma_{b}\boldsymbol{1}||^{\alpha}\delta_{\frac{\boldsymbol{1}}{||\boldsymbol{1}||}}+\int||\sigma_{w}(\phi\circ{f})||^{\alpha}\delta_{\frac{\phi\circ{f}}{||\phi\circ{f}||}}q^{(l-1)}(\text{d}{f}).
\end{equation}

\newpage

\section{Finite-dimensional projections} \label{sec:fd_proj}

We show that, as $n\rightarrow+\infty$,
\begin{equation}
({f}_{i}^{(l)}(\boldsymbol{X},n))_{i\geq1}\overset{w}{\longrightarrow}\bigotimes_{i\geq1}\text{St}_{k}(\alpha,\Gamma(l)),
\end{equation}
by proving the large $n$ asymptotic behavior of any finite linear combination of the ${f}_{i}^{(l)}(\boldsymbol{X},n)$'s, for $i\in\mathcal{L}\subset\mathbb{N}$. See, e.g. \cite{billingsley1999convergence} and reference therein. Following the notation of \cite{matthews2018gaussian}, consider a finite linear combination of the function values without the bias. In other terms, let us consider
\begin{equation*}
T^{(l)}(\mathcal{L},p,\boldsymbol{X},n)=\sum_{i\in \mathcal{L}}p_{i}[{f}_{i}^{(l)}(\boldsymbol{X},n)-b_{i}^{(l)}\boldsymbol{1}].
\end{equation*}
Then, we write
\begin{align*}
T^{(l)}(\mathcal{L},p,\boldsymbol{X},n)&=\sum_{i\in \mathcal{L}}p_{i}[{f}_{i}^{(l)}(\boldsymbol{X},n)-b_{i}^{(l)}\boldsymbol{1}]\\
&=\sum_{i\in \mathcal{L}}p_{i}\left[\frac{1}{n^{1/\alpha}}\sum_{j=1}^{n}w_{i,j}^{(l)}(\phi\circ {f}_{j}^{(l-1)}(\boldsymbol{X},n))\right]\\
&=\frac{1}{n^{1/\alpha}}\sum_{j=1}^{n}\sum_{i\in \mathcal{L}}p_{i}w_{i,j}^{(l)}(\phi\circ{f}_{j}^{(l-1)}(\boldsymbol{X},n))\\
&=\frac{1}{n^{1/\alpha}}\sum_{j=1}^{n}\gamma_{j}^{(l)}(\mathcal{L},p,\boldsymbol{X},n),
\end{align*}
where
\begin{equation*}
\gamma_{j}^{(l)}(\mathcal{L},p,\boldsymbol{X},n)=\sum_{i\in \mathcal{L}}p_{i}w_{i,j}^{(l)}(\phi\circ{f}_{j}^{(l-1)}(\boldsymbol{X},n)).
\end{equation*}
Then,
\begin{align*}
&\varphi_{T^{(l)}(\mathcal{L},p,\mathbf{X},n)\,|\,\{{f}_{j}^{(l-1)}(\boldsymbol{X},n)\}_{j=1,\ldots,n}}(\boldsymbol{t})\\
&\quad=\mathbb{E}[e^{\textrm{i}\boldsymbol{t}^{T}T^{(l)}(\mathcal{L},p,\boldsymbol{X},n)}\,|\,\{{f}_{j}^{(l-1)}(\boldsymbol{X},n)\}_{j=1,\ldots,n}]\\
&\quad=\mathbb{E}\left[\exp\left\{\textrm{i}\boldsymbol{t}^{T}\left[\frac{1}{n^{1/\alpha}}\sum_{j=1}^{n}\sum_{i\in \mathcal{L}}\alpha_{i}w_{i,j}^{(l)}(\phi\circ{f}_{j}^{(l-1)}(\boldsymbol{X},n))\right]\right\}\,|\,\{{f}_{j}^{(l-1)}(\boldsymbol{X},n)\}_{j=1,\ldots,n}\right]\\
&\quad=\prod_{j=1}^{n}\prod_{i\in\mathcal{L}}\mathbb{E}\left[\exp\left\{\textrm{i}\boldsymbol{t}^{T}\frac{1}{n^{1/\alpha}}p_{i}w_{i,j}^{(l)}(\phi\circ{f}_{j}^{(l-1)}(\boldsymbol{X},n))\,|\,\{{f}_{j}^{(l-1)}(\boldsymbol{X},n)\}_{j=1,\ldots,n}\right\}\right]\\
&\quad=\prod_{j=1}^{n}\prod_{i\in\mathcal{L}}\text{e}^{-\frac{p^{\alpha}_{i}\sigma_{w}^{\alpha}}{n}|\boldsymbol{t}^{T}(\phi\circ f_{j}^{(l-1)}(\boldsymbol{X},n))|^{\alpha}}\\
&\quad=\exp\left\{-\int_{\mathbb{S}^{k-1}}|\boldsymbol{t}^{T}\boldsymbol{s}|^{\alpha}\left(\frac{1}{n}\sum_{j=1}^{n}\sum_{i\in\mathcal{L}}||p_{i}\sigma_{w}(\phi\circ f_{j}^{(l-1)}(\boldsymbol{X},n))||^{\alpha}\delta_{\frac{\phi\circ f_{j}^{(l-1)}(\boldsymbol{X},n)}{||\phi\circ f_{j}^{(l-1)}(\boldsymbol{X},n)||}}\right)(\text{d}\boldsymbol{s})\right\}
\end{align*}
That is, 
\begin{align*}
&T^{(l)}(\mathcal{L},p,\boldsymbol{X},n)\,|\,\{{f}_{j}^{(l-1)}(\boldsymbol{X},n)\}_{j=1,\ldots,n}\overset{\text{d}}{=}\boldsymbol{S}_{\alpha,\Gamma^{(l)}}.
\end{align*}
where
\begin{equation*}
\Gamma^{(l)}=\frac{1}{n}\sum_{j=1}^{n}\sum_{i\in\mathcal{L}}||p_{i}\sigma_{w}(\phi\circ f_{j}^{(l-1)}(\boldsymbol{X},n))||^{\alpha}\delta_{\frac{\phi\circ f_{j}^{(l-1)}(\boldsymbol{X},n)}{||\phi\circ f_{j}^{(l-1)}(\boldsymbol{X},n)||}}.
\end{equation*}
Then, along lines similar to the proof of the large $n$ asymptotics for the $i$-th coordinate, we have
\begin{align*}
&\mathbb{E}[\text{e}^{\textrm{i}\boldsymbol{t}^{T}T^{(l)}(\mathcal{L},p,\boldsymbol{X},n)}]\\
&\quad\rightarrow \exp\left\{-\int\int_{\mathbb{S}^{k-1}}|\boldsymbol{t}^{T}\boldsymbol{s}|^{\alpha}\left(\sum_{i\in\mathcal{L}}||p_{i}\sigma_{w}(\phi\circ{f})||^{\alpha}\delta_{\frac{\phi\circ{f}}{||\phi\circ{f}||}}\right)(\text{d}\boldsymbol{s})q^{(l-1)}(\text{d}{f})\right\}
\end{align*}
as $n\rightarrow+\infty$. This complete the proof.

\newpage

\section{Numerical evaluation of the recursion}

\begin{figure}
    \centering
    \hspace*{-0.5cm}\includegraphics[width=\linewidth]{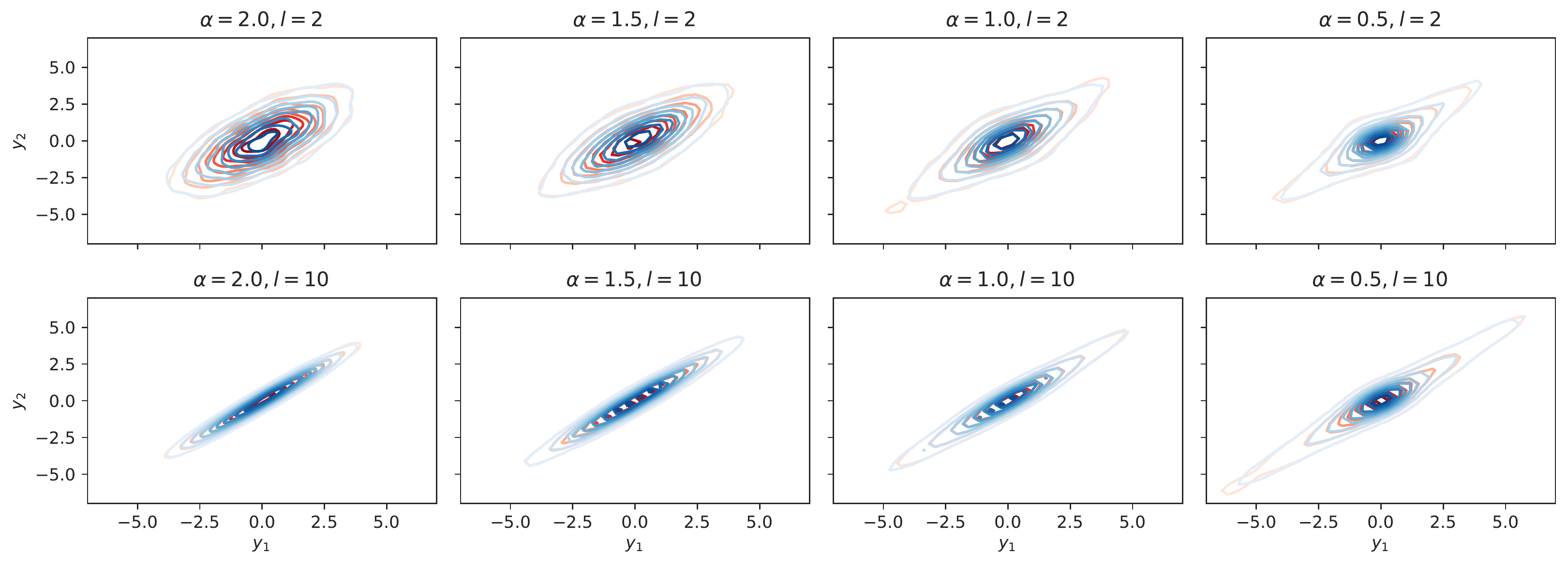}
    \caption{2D-KDE estimates for $y \sim f^{(l)}_1(x,x',n)$ (red) and $y \sim \text{St}_k(\alpha,\widetilde{\Gamma}(x,x',l,M))$ (blue).}
    \label{fig:kde_recursion}
\end{figure}

In this section we perform a preliminary numerical investigation of the approach proposed in \cref{sec:bayesian_inference} for the evaluation of recursion \cref{eq:recursion_1}-\cref{eq:recursion_l}. We consider only the case of two inputs $x=-0.5,x'=1.0$ (i.e. a bivariate stable distribution) and we use pseudo random numbers, i.e. standard Monte Carlo (MC), instead of quasi random numbers as suggested in the main text. We consider $\sigma_b = \sigma_w = 1$, the tanh activation function, different values of the stability index $\alpha$, and both shallow ($l=2$, i.e. $1$ hidden layer) and deep ($l=10$) NNs. In all cases the networks are wide: $n=300$. In \cref{fig:kde_recursion} we compare the bivariate distributions of: i) the first dimension ($i=1$) of the NN distribution $y \sim f^{(l)}_1(x,x',n)$ ii) its asymptotic distribution $y \sim \text{St}_k(\alpha,\widetilde{\Gamma}(x,x',l,M))$ as $n \rightarrow +\infty$. In ii) we use $M=1000$ MC samples to evaluate the discrete spectral measure $\widetilde{\Gamma}$ at each layer. In both i) and ii) we generate $100.000$ samples for $y \in \mathbb{R}^2$ that are used to obtain the 2D-KDE plots of \cref{fig:kde_recursion}. We can observe close agreement in all cases considered (the "squarish" level curves near the central regions for small $\alpha$ are an artifact due to the specific KDE estimation algorithm employed and its non-robustness to "outliers").

The code at \url{https://github.com/stepelu/deep-stable} contains a \texttt{numpy}-based Python implementation of the algorithms used for the simulation of scalar and multivariate stable distributions. Scalar stable variables are generated according to \cite{weron1996chambers,weron2010correction}. In the case of multivariate stable variables the algorithm implemented is the one reported in \cite{nolan2008overview}, note that the discrete spectral measure needs to be symmetrized. The code also contains the routines used to sample from $f^{(l)}(x,n)$ and from $\text{St}_k(\alpha,\widetilde{\Gamma}(x,l,M))$. The implementation does not rely on advanced features so it is easily portable to deep learning frameworks such as \texttt{tensorflow} or \texttt{pytorch}. By modifying the calls to uniform random generators, it is also possible to use quasi random number generators. In all cases, the usual precaution to exclude the extremes of the supports of the uniform distributions involved (i.e. to sample from $\mathcal{U}(0,1)$, not from $\mathcal{U}[0,1]$) applies.

\newpage

\section{Empirical analysis of trained CNNs}

In this section we investigate whether trained models exhibit parameters distributions close to that of Stable distributions with stability index $0 < \alpha < 2$, i.e. non-Gaussian. We consider 3 models from the PyTorch's TorchVision repository, i.e. CNNs trained on ImageNet. While fully connected networks are ideal starting points for a theoretical analysis, it seems possible to expand our results to CNNs as done in \cite{garriga-alonso2018deep} for Gaussian Processes (GP). This allows us to investigate the parameter distributions of trained model in the "realistic" setting of overparametrized models applied to big datasets with the use of batch normalization and adaptive optimizers. 

\begin{figure}
    \centering
    \hspace*{-0.5cm}\includegraphics[width=0.8\linewidth]{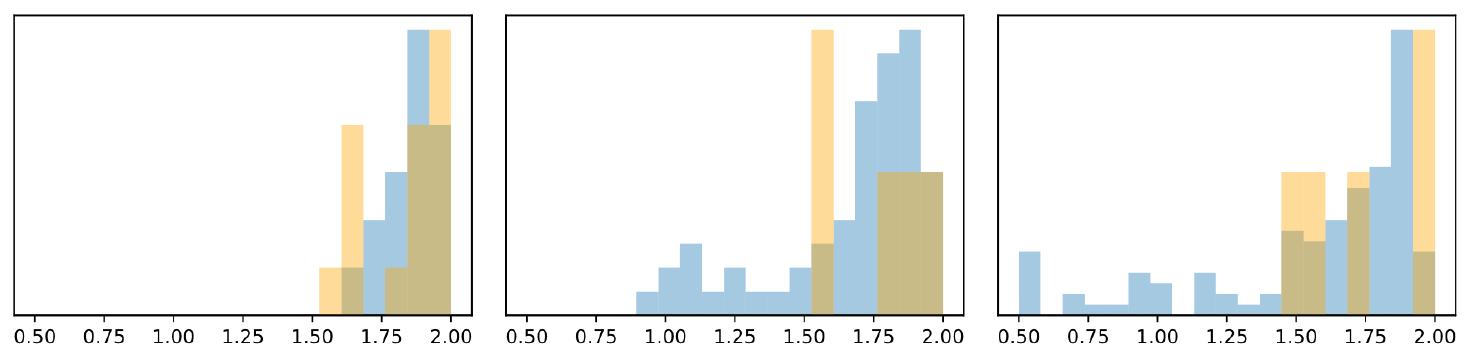}
    \caption{Histograms for the stability indexes $\alpha$ of the fitted Stable distributions for all layers, in blue for the weights and in yellow for the biases. The models, from left-to-right, are: VGG-16, ResNet-50 and ResNet-101.}
    \label{fig:index_histogram}
\end{figure}

We restrict our analysis to marginal distributions and for each layer we collect all weights (CNN filters) and biases and fit a Stable distribution via maximum likelihood estimation (MLE). In \cref{fig:index_histogram} we plot histograms for the stability indexes $\alpha$ of the fitted Stable distributions for all layers. We see that distributions are often non-Gaussian, and $\alpha$ seems to be decreasing with the depth of the model. However, it is not possible to draw definitive conclusions from this short experiment.

\begin{figure}
    \centering
    \hspace*{-0.5cm}\includegraphics[width=0.8\linewidth]{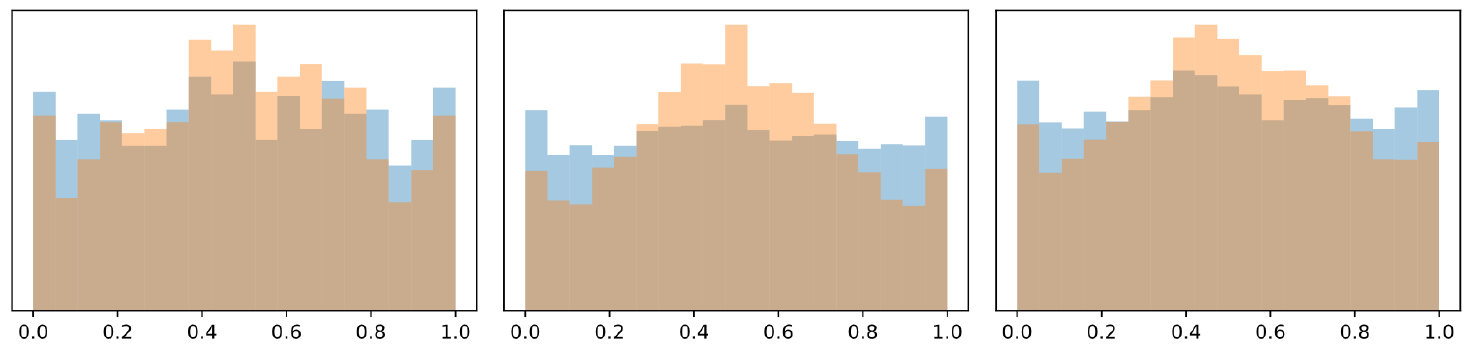}
    \caption{Histograms of the cdf evaluations for the Stable distribution (blue) and Gaussian distribution (orange) fitted to the weights of the first three layers (left-to-right) of the VGG16 model.}
    \label{fig:cdf_histogram}
\end{figure}

To obtain an indication of the goodness of fit of Stable distributions to the parameters, for the first three layers of VGG-16 ($\alpha \sim 1.7$) we: fit a Stable distribution to the weights; compute the cumulative distribution function (cdf) of this Stable distribution for each weight; fit a Gaussian distribution to the weights; compute the cdf of this Gaussian distribution for each weight; plot in \cref{fig:cdf_histogram} a histogram of the cdf evaluations for the Stable and Gaussian distribution. In case of perfect fit the histogram should be flat, as the cdf evaluations should be iid uniformly distributed. We see that the fit of the Stable distributions is as expected better than the fit of Gaussian distributions, especially in the tails. The peculiar behavior at extremes of the histograms (tails) could be due to the use of truncated initializations in PyTorch. We validated MLE (limited here to $\alpha > 0.5$) and cdf computation on synthetic data generated via $\texttt{sample\_stable()}$ from the code accompanying this paper.

\end{document}